\documentclass[lettersize,journal]{IEEEtran}
\usepackage{amsmath,amsfonts}
\usepackage{algorithmic}
\usepackage{algorithm}
\usepackage{array}
\usepackage[caption=false,font=normalsize,labelfont=sf,textfont=sf]{subfig}
\usepackage{textcomp}
\usepackage{stfloats}
\usepackage{url}
\usepackage{verbatim}
\usepackage{graphicx}
\usepackage{cite}
\usepackage{colortbl}
\begin{document}

\title{TextFusion: Unveiling the Power of Textual Semantics for Controllable Image Fusion}

\author{Chunyang Cheng,
        Tianyang Xu,
        Xiao-Jun Wu,
        Hui Li,
        Xi Li,
        Zhangyong Tang,
        and Josef Kittler
\thanks{C. Cheng, T. Xu, X.-J. Wu, H. Li and Z. Tang are with the School of Artificial Intelligence and Computer Science, Jiangnan University, Wuxi 214122, P.R. China. (Corresponding author: X.-J. Wu, e-mail: wu\_xiaojun@jiangnan.edu.cn)  }
\thanks{Xi Li is with the College of Computer Science and Technology, Zhejiang
University, Hangzhou 310027, China. (e-mail: xilizju@zju.edu.cn)}
\thanks{Josef Kittler is with the Centre for Vision, Speech and Signal Processing, University of Surrey, Guildford GU2 7XH, U.K. (e-mail: j.kittler@surrey.ac.uk)}}

\markboth{Journal of \LaTeX\ Class Files,~Vol.~0, No.~0, FEB~2024}%
{Shell \MakeLowercase{\textit{et al.}}: A Sample Article Using IEEEtran.cls for IEEE Journals}


\maketitle

\begin{abstract}
Advanced image fusion methods are devoted to generating the fusion results by aggregating the complementary information conveyed by the source images.
However, the difference in the source-specific manifestation of the imaged scene content makes it difficult to design a robust and controllable fusion process.
We argue that this issue can be alleviated with the help of higher-level semantics, conveyed by the text modality, which should enable us to generate fused images for different purposes, such as visualisation and downstream tasks, in a controllable way.
This is achieved by exploiting a vision-and-language model to build a coarse-to-fine association mechanism between the text and image signals.
With the guidance of the association maps, an affine fusion unit is embedded in the transformer network to fuse the text and vision modalities at the feature level.
As another ingredient of this work, we propose the use of textual attention to adapt image quality assessment to the fusion task.
To facilitate the implementation of the proposed text-guided fusion paradigm, and its adoption by the wider research community, we release a text-annotated image fusion dataset IVT.
Extensive experiments demonstrate that our approach (TextFusion) consistently outperforms traditional appearance-based fusion methods.
Our code and dataset will be publicly available at \url{https://github.com/AWCXV/TextFusion}.
\end{abstract}


\section{Introduction}
\label{sec:intro}
\IEEEPARstart{I}{mage} fusion is a technique allowing to combine multiple input images {acquired by different sensors or shooting configurations}, into a single robust fused image which contains more information.
The output image should be closely aligned with human perception, with the salient information from the source being well preserved~\cite{zhang2023IVIFsurveyPAMI,hermessi2021multimodalMedical,zhang2021multisurveypami, xu2022mefsurvey}.
As an illustration drawn from the thermal infrared and visible (RGB) image fusion paradigm, this fusion task demands a comprehensive preservation of information from both inputs, \textit{i.e.}, the thermal radiation information derived from the infrared modality and the prominent texture detail from the visible image.
Besides ameliorating the visual perception of the multimodal data, the fusion results are expected to boost the performance of downstream computer vision tasks~\cite{karim2023fusionSurveyInfFus,xiao2022rgbtTrackingAAAI}.

\begin{figure}[t]
\centering
\includegraphics[width=1\linewidth]{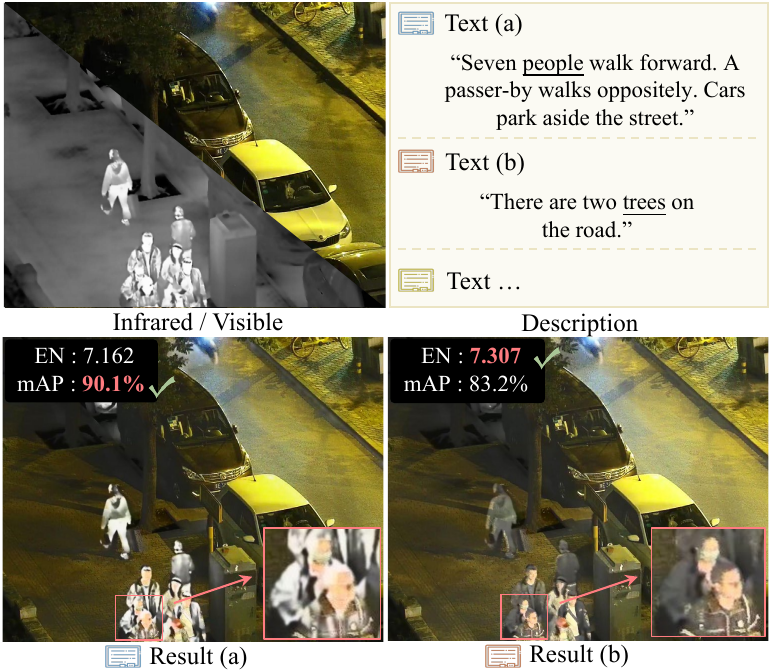}
\caption{The input modalities and two fusion results obtained by our TextFusion with different description. As reflected in the metrics, each application scenario requires a different fusion scheme to achieve the best performance. (EN is the image quality metric of information entropy and mAP denotes the mean average detection precision.)
}
\label{figure_fusionPurposes}
\end{figure}

\begin{figure}[t]
\centering
\includegraphics[width=\linewidth]{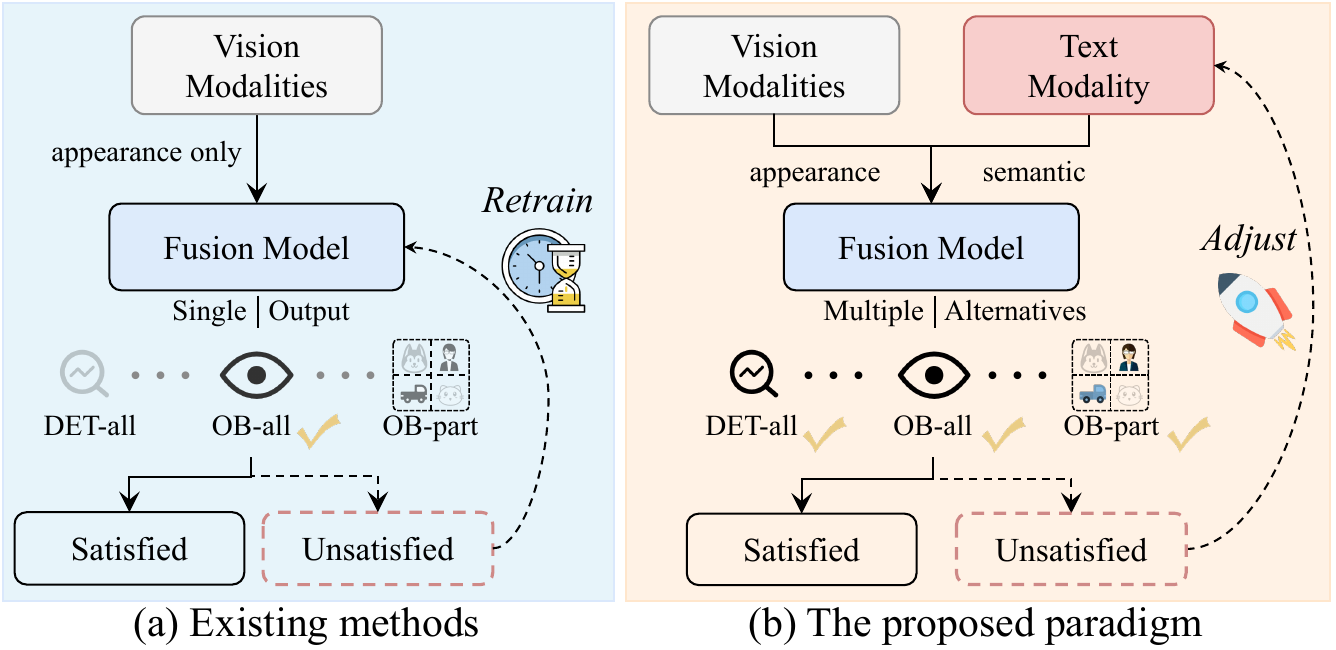}
\caption{Existing learning-based image fusion methods and the proposed controllable image fusion paradigm.
To generate appropriate fusion results for a specific scenario (different tasks or concerned objects), existing methods cannot realise it or require expensive retraining. The same goal can be achieved by simply adjusting the focused objectives of textual description in our paradigm. (DET-all: general detection; OB-all: observation for the whole scene; OB-part: observation for the interested regions)
}
\label{figure_comparison_existing_proposed}
\end{figure}

The effectiveness of an image fusion approach will depend on the application scenes, \textit{i.e.}, whether it relates to,~\textit{e.g.}, observation of the overall scene or area of interest, detection, or recognition.
As shown in Fig.~\ref{figure_fusionPurposes},
when image fusion is used in a detection task, the result (a) obtained by our method might be favoured, since it takes advantage of certain thermal radiation information to enhance the image contrast.
On the other hand, result (b) is probably more appropriate for human observation, as it conveys more information (higher EN) and the image is rich in texture details (yellow boxes).
However, such dependencies have not been studied in the existing literature.
Instead, by relying purely on visual information, once trained, the current approaches are capable of producing fused images only in a fixed manner.
Consequently, retraining fusion systems is required for specific scenario or concerned objects (Fig.~\ref{figure_comparison_existing_proposed} (a)), which limits their practical utility.

In this paper, we address the aforementioned \textit{lack of controllability} by offering a variety of potential fusion style alternatives.
Specifically, our approach is to provide control over the fusion process using a textual description for the input vision modalities, which focuses on the fusion objectives.
The text modality contains higher-level semantics for the image fusion task, overcoming the unwarranted reliance purely on the vision modalities.
As shown in Fig.~\ref{figure_comparison_existing_proposed}~(b), the proposed paradigm is able to deliver diverse image fusion outputs for different scenes using a single image fusion model.
A change of the fusion focus (concerned objects or regions) can be realised by simply adjusting the textual description for the input vision modalities.
In general, our innovation introduces the use of linguistic information in the image fusion field.

The adoption of the text modality into the image fusion field raises significant challenges.
The first problem is posed by the need to perform a simultaneous fusion of features computed by three different modalities. This is very different from the existing tasks that typically involve only the RGB and text modalities.
The second issue is raised by the huge semantic gap between the high-level linguistic input and the appearance information in the vision signals. This gap impedes the text modality in boosting the fine-grained fusion task.
Thirdly, a challenge also arises from the lack of available image fusion datasets that include text annotations.

To handle the first challenge, we introduce an affine fusion unit as the foundational component of the TextFusion model.
This unit is designed to fuse the vision features based on the guidance provided by the linguistic information.
Secondly, to address the semantic gap, a coarse-to-fine text-vision association mechanism is proposed to produce a pixel-level relation map.
This link (relation map) enables the textual information to be injected into the optimization process.
Regarding the third issue, we curate and release a new IVT dataset that contains $11,450$ aligned infrared and visible image pairs together with manually annotated textual descriptions.

Another important consideration is that the commonly used image fusion metrics emphasise the fusion result quality assessment for tasks with ground truth (GT) output, \textit{e.g.}, image super-resolution and dehazing~\cite{chao2023equivalent,song2023visionDehaze}.
Certain metrics simply employ averaging to calculate the contribution of the source images, without incorporating any adaptations tailored to the fusion task.
These metrics are not fit for the purpose of task-specific fusion. We argue that it is essential to replace the traditional indices with the proposed textual attention metrics to rectify this problem.
In particular, we propose to gauge the fusion image quality by focusing on regions corresponding to the semantic preference expressed by the text prompt.

The contribution of our work can be summarized as follows:
\begin{itemize}
\item We conduct \textbf{a pioneering investigation} of the merits of controlling image fusion by textual specification, which defines the semantic content of interest in the source images. This constitutes a new paradigm in the image fusion area.
\item  We propose a new image fusion model based on an affine fusion scheme that automatically adapts to the semantic image content defined by the given textual input. 
\item To facilitate the multimodal data study (image and language), we release a new dataset containing different textual descriptions for RGBT image pairs.
\item We introduce textual attention-dependent image quality measures. 
They reflect differences in the emphasis of the fusion process on the salient image foreground specified linguistically, and the complementary image background.
\end{itemize}

\section{Related work}
\label{sec:related}
\subsection{Learning-based image fusion methods}
Deep image fusion methods can be divided into two categories, according to the training paradigms.
In the first group, a convolutional neural network (CNN) is used to obtain 
features of the source images, or intermediate outputs~\cite{2017Multi,li2018densefuse,2021zhaoDepth_distill_mf,li2021rfn,zhao2023cddfuse}.
Extra steps, \textit{e.g.}, filtering techniques, fusion rules design, and other post-processing are required to generate the final output.
These operations are conducted on the entire source images without considering any specific image content, thereby increasing the difficulty of obtaining fine-grained fusion results.

In contrast, there are studies that are trying to use CNNs to generate the fused images directly, formulating the fusion task in an end-to-end manner~\cite{zhang2021sdnet,fu2021fuimageGAN,zhang2020ifcnn,rao2022tgfuse,cheng2023mufusion,xu2023murf}.
To define effective supervision signals, loss functions are designed to impose different image quality-related constraints on the relationship between the input and output images~\cite{zhao2020realmAdaption,he2023degradationICCV2023}.
Without any semantic guidance, the loss function design shares similar risks with the aforementioned handcrafted fusion rules.
Currently, scholars attempt to address this issue by jointly training the fusion network and downstream vision models~\cite{tang2022SeAFusion,liu2022target,TANG2023divfusion,Zhao2023metafusion,liu2023fusion_seg_ICCV2023}.
Nevertheless, the fulfillment of diverse application requirements by these approaches necessitates the training of multiple models.
In other words, the approach has no mechanism to control the fused image, and therefore lacks flexibility.

\subsection{Vision-and-language pre-trained models}
In recent years, vision-and-language pre-trained models have received wide attention~\cite{gan2020vlp1,zhang2021vlp2,zhou2022unsupervisedvilt}.
These models are trained on large datasets, exhibiting extraordinary expressive power and generalisation ability, which make them capable of boosting the performance of downstream joint vision-and-language subtasks~\cite{kim2021vilt},~\textit{e.g.}, retrieval, image caption, and visual question answering.
However, their applicability to the image fusion task is an open problem.
In other words, the merits brought by the textual semantics to computer vision via vision-and-language processing have not been explored in advanced fusion approaches.

\subsection{Text prompt learning}
\label{section_relatedTextPrompt}
Text information has been studied in many vision tasks, formulating a text prompt learning paradigm.
In ManiGAN, Li~\textit{et.al.} successfully realised image manipulation based on the textual description guidance~\cite{li2020manigan}.
Note that, the target images only include some simple scenes, \textit{e.g.}, a photo of a bird. 
Similar attempts are also made in some high-level vision tasks, \textit{e.g.}, segmentation~\cite{li2022languageSegmentation,luddecke2022textImageSegmentation,lin2023clipSeg} and tracking~\cite{li2023ovtrack}.
Nevertheless, due to the additional vision modality (infrared) here, the introduction of the text modality in the image fusion task is much more challenging.
Specifically, related conclusions and operations that only involve the RGB information cannot be directly used in the fusion research.

\begin{figure}[t]
\centering
\includegraphics[width=1\linewidth]{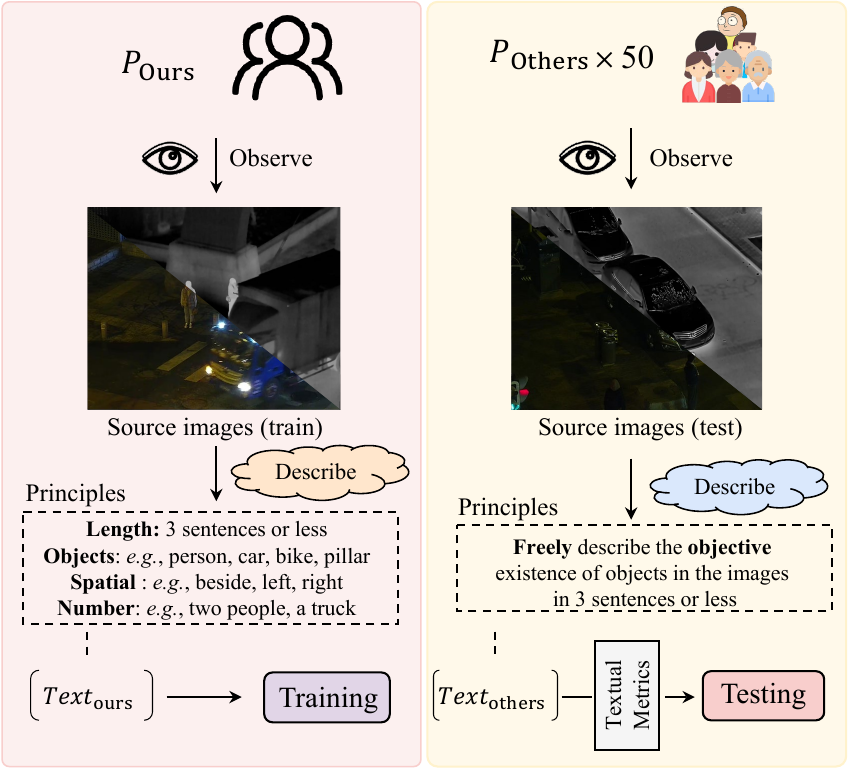}
\caption{
An illustration of different annotation manners in the training and testing phases.
Considering different annotation principles, our research group and independent volunteers are the observers of different stages in the text-guided image fusion paradigm, respectively, to avoid potential bias issues.}
\label{figure_annotation_differences}
\end{figure}

\begin{figure}[t]
\centering
\includegraphics[width=1\linewidth]{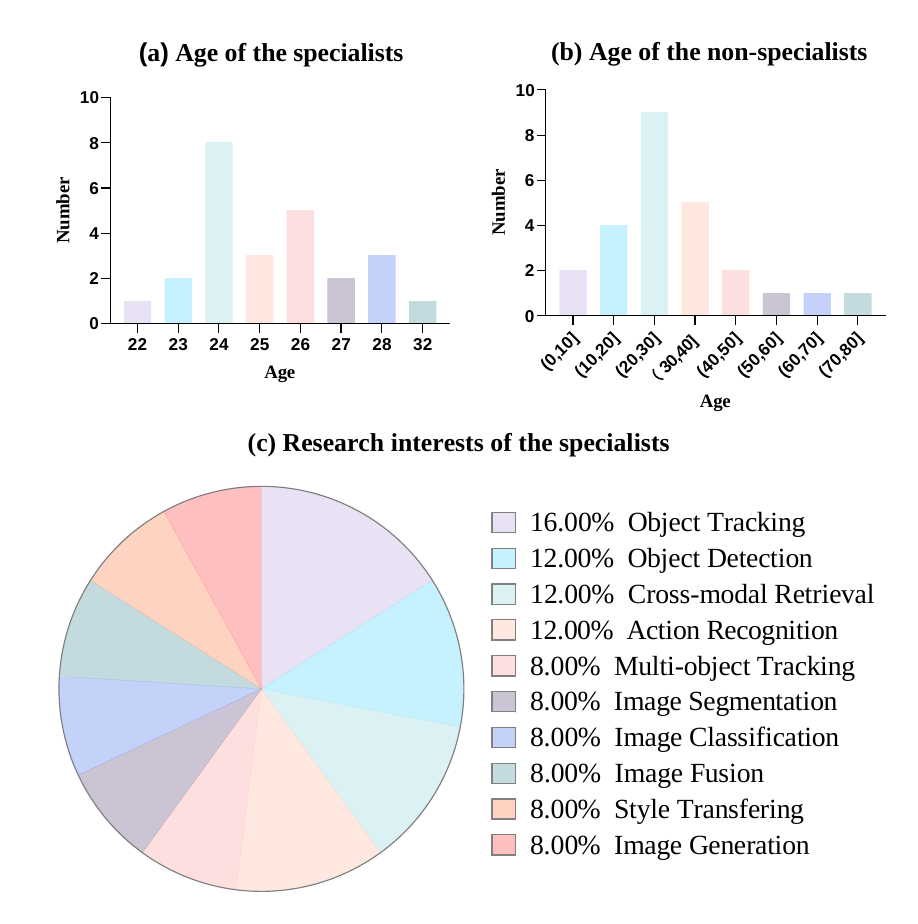}
\caption{
The demographic information of the annotation volunteers.
(a) and (c) denote the age and research interest information of the specialists. (b) denotes the age information of the non-specialists.
As shown in the charts, the observers range from different ages and areas, which can represent a general view of human beings for understanding the RGBT image pairs.}

\label{figure_observers_statistics}
\end{figure}

\section{The IVT dataset}
\label{moreDetailsIVT}
In this paper, we create a new multi-modal dataset which consists of aligned infrared and visible image pairs and corresponding textual descriptions (IVT dataset).
In this section, we will provide more details about this dataset.
As shown in Fig.~\ref{figure_annotation_differences}, we first present the annotation differences in the training and testing processes.
Specifically, in the training stage, our research group ($P_{\mathrm{ours}}$)  describes the image pairs according to several principles, \textit{e.g.}, focusing on the description of the objects and their spatial information in the scenes.
These text annotations will be directly used to train the fusion network.

On the other hand, to mitigate the bias issues in the testing phase, we expect the textual description to cover the perspectives of various persons.
Therefore, we invite some volunteers ($P_{\mathrm{others}}$) to freely structure the words and describe the objective existence of different targets in three sentences or less.
Our research group was not involved in any of this annotation process.
Finally, using the annotated data, we generate more descriptions of different lengths for each image pair by concatenating the subsets of the sentences.

More specifically, for the testing dataset, 50 irrelevant people are involved in the annotation process.
In Fig.~\ref{figure_observers_statistics}, we summarize the demographic information of these persons and divide them into two categories: specialists and non-specialists.
In our setting, there are 25 people in each category.
For the specialists, as shown in Fig.~\ref{figure_observers_statistics} (a) and (c), they are all teenagers ranging from 20 to 30 years old and their research interests cover different areas of computer vision tasks or image processing techniques.
Meanwhile, as seen from Fig.~\ref{figure_observers_statistics} (b), we also ask some non-specialist people from children under 10 years old to 70-year-old seniors to participate in the annotation process as supplements for the experts' opinions.
In this way, combining the comprehensive descriptions from different persons and the textual attention mechanism, we can better evaluate the fusion results with the guidance of the text modality.

\section{Approach}
\label{sec:approach}
\label{method}

In our TextFusion, we receive two vision modalities as input and fuse them under the guidance of the text modality.
As an illustration, the input visual signals are infrared $I_{\mathrm{ir}}$ and visible $I_{\mathrm{vis}}$ images.
The fusion result is denoted as $I_{\mathrm{f}}$.

\begin{figure*}[t]
\centering
\includegraphics[width=1\linewidth]{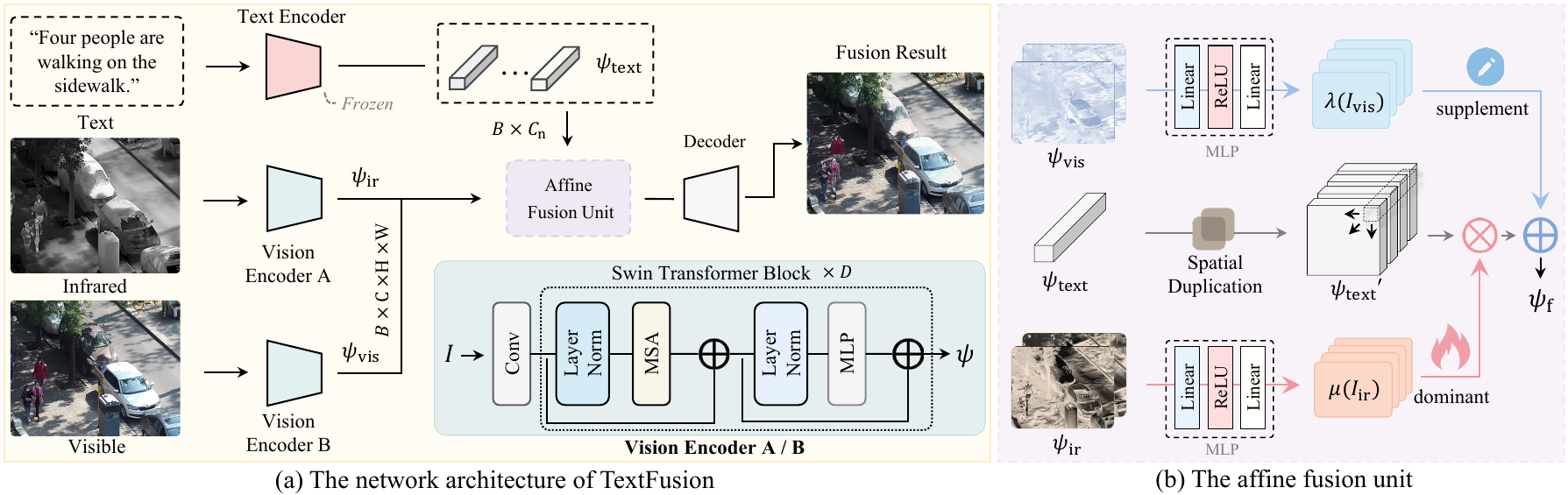}
\caption{An illustration of the TextFusion model and the affine fusion unit design.
(a): Our fusion model receives the input image pairs and the textual description as input.
The text encoder from the CLIP model and two vision encoders based on the Swin Transformer Blocks are used to extract the hidden representations of source input.
We further propose an affine fusion model to align and aggregate these features.
Subsequently, we use a decoder consisting of convolutional layers to reconstruct the fused image.
(b): The affine fusion unit is used to fuse the vision signals with the help of the text modality.
In our design, the infrared modality is used to generate the weight term $\mu$, while the bias term $\lambda$ is calculated based on the visible images.
We expand the spatial dimension of the text features to match the weight term prior to performing the element-wise multiplication operation.
}
\label{figure_networkArchitecture}
\end{figure*}

\subsection{Network architecture and the affine fusion unit}
As shown in Fig.~\ref{figure_networkArchitecture} (a), our network is composed of three encoders, \textit{i.e.}, two vision encoders and a text encoder.
For the text encoder, we use the popular CLIP model to obtain linguistic features $\psi_{\mathrm{text}}$.
On the other hand, the vision encoder part of CLIP contains down-sampling operations, which will impact the image quality when restoring the original size.
Thus, to maintain the performance and keep computation efficient, we further design two vision encoders based on the Swin Transformer Blocks~\cite{Liu_2021SwinTransformer} for extracting image features $\psi_{\mathrm{ir}}$ and $\psi_{\mathrm{vis}}$.
After that, we insert an affine fusion unit into the data processing chain to integrate the extracted features.
The final fused image is then reconstructed by a decoder, consisting of convolutional layers.

Alternatively, a straightforward approach for fusing the features of these three modalities would be to perform the concatenation operation along the channel dimension.
Unfortunately, such a coarse-grain operation precludes the network from learning the correspondence between the focus of interest defined by the textual description and the visual content.
Inspired by some text-and-vision tasks~\cite{li2022languageSegmentation,li2020manigan}, we propose an affine fusion unit to address this issue.

In particular, our intention is to utilize the text information to exert control over the fusion results. One of the criteria that distinguishes different fusion scheme behaviours is the degree of preservation of the thermal radiation content.
As shown in Fig.~\ref{figure_networkArchitecture} (b), following the theory in~\cite{liu2020bilevel}, in our approach, we assume that the infrared image is the dominant modality for producing the fusion results, and the weight term of the affine transformation is generated accordingly.
The text features in the proposed approach are used to identify the salient attributes of the source input specified by the prompt.
The bias term, on the other hand, is derived from the visible image as we consider this aspect to be text-independent and solely used for reconstructing a high-quality fusion result.

Mathematically, the information fusion process of this unit can be formulated as:
\begin{equation}
    \psi_{f} = \mu(I_{\mathrm{ir}})\cdot \psi_{\mathrm{text}}^{'} + \lambda(\mathrm{I_{vis}}),
\end{equation}
where $\mu$ and $\lambda$ denote the learned weight and bias terms.
$\mu(I_{\mathrm{ir}})$ and $\lambda(\mathrm{I_{vis}})$ are generated by a multi-layer perceptron structure (MLP).
The purpose of using MLP is to augment the channels of the original image features to match those of the text features.
Simultaneously, we perform duplication operations to expand the spatial dimension of the linguistic feature vector, aligning it with the resolution of the visual features.

\begin{figure}[t]
\centering
\includegraphics[width=1\linewidth]{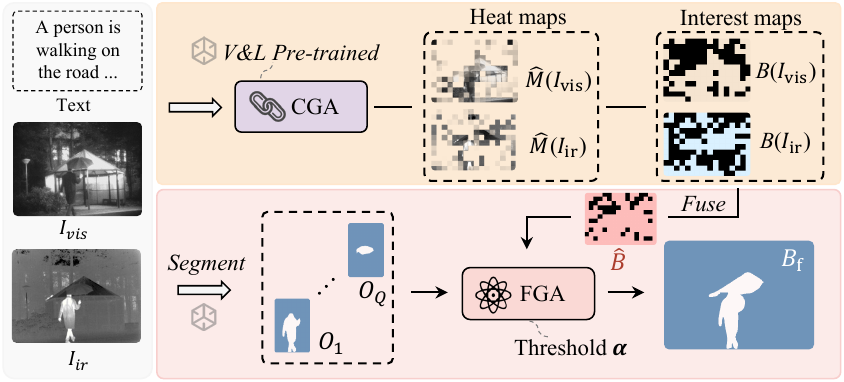}
\caption{Pipeline of the coarse-to-fine association mechanism. 
CGA and FGA denote the coarse-grained and fined-grained association modules, respectively.
The input images together with textual description are first fed into the vision and languange pre-trained model model to obtain the cross-modal heat maps $M_{\mathrm{w}_i}$.
Then, we aggregate heat maps of the nouns into a single one $\hat{M}$.
This heat map will be further used to generate the binary interested map $B$, with subsequent maximizing operation used to yield the initial interested map $\hat{B}$.
Finally, we fine-tune this map based on the segmentation techniques and a threshold $\alpha$ in the FGA to produce the final interested map $B_{\mathrm{f}}$.
}
\label{figure_generateFinetunedSegmentationResults}
\end{figure}

\subsection{Association and training processes}
\subsubsection{Coarse-to-fine association}
As there is no GT image in the fusion task, to control the fused image via linguistic guidance, we need to embed the textual semantics into the loss function design.
Thus, we need to generate an association map between the modalities for the training process.
However, obtaining a pixel-wise relation map between vision and text modalities can be very expensive~\cite{huang2020pixelBert}.
Thus, a coarse-to-fine association mechanism is proposed to alleviate this challenge.

As shown in Fig.~\ref{figure_generateFinetunedSegmentationResults}, aiming to distinguish the user-defined content of interest in the input images from the background, we first construct a coarse-grained association (CGA) module to obtain the heat map $M$ linking each word to visual content, \textit{i.e.}, identifying the degree of correlation between the words and image regions.
The heat maps are produced by a vision-and-language pre-trained model.
For each noun $\mathrm{w}_j$, we fuse their heat maps into a single one $\hat{M}(I)$ by taking the maximum value at each location of these heat maps, \textit{i.e.}
\begin{equation}
    \hat{M}(I)(x,y) = \max\{M_{\mathrm{w}_j}(I)(x,y)\}. 
\end{equation}

These nouns indicate the existence of such objects in the scene, and identify the salient regions, which should be considered with priority in the fused image. We record these areas by using the binary interest maps $B({I_{\mathrm{vis}}})$ and $B(I_{\mathrm{ir}})$.
To fully utilize the information from both images, we also aggregate these two interest maps into a single one $\hat{B}$ based on the choose-max strategy.

Note that, the vision-and-language pre-trained model used in our CGA takes input and produces output in a patch-wise format~\cite{kim2021vilt}.
This patch-wise representation may result in the loss of edge contours of instances, making the use of the generated patches in the training process challenging.

Taking this into consideration, we propose a subsequent fine-grained association (FGA) module to fine-tune the interest map.
Such association is determined based on $Q$ instances of source images,~\textit{e.g.}, persons, trees and cars, from a segmentation model. 
We first calculate the overlapping ratio $r_i$ between these instances and the interest map,
\textit{i.e.}
\begin{equation}
    r_i=\frac{\sum_{x=1}^H\sum_{y=1}^W(O_i(x,y)\cdot \hat{B}(x,y))}{\sum_{x=1}^H\sum_{y=1}^WO_i(x,y)},
\end{equation}
where $i\in\{1,2,...,Q\}$ , $H$ and $W$ are the height and width of the input image, respectively.

\begin{figure*}[t]
\centering
\includegraphics[width=0.8\linewidth]{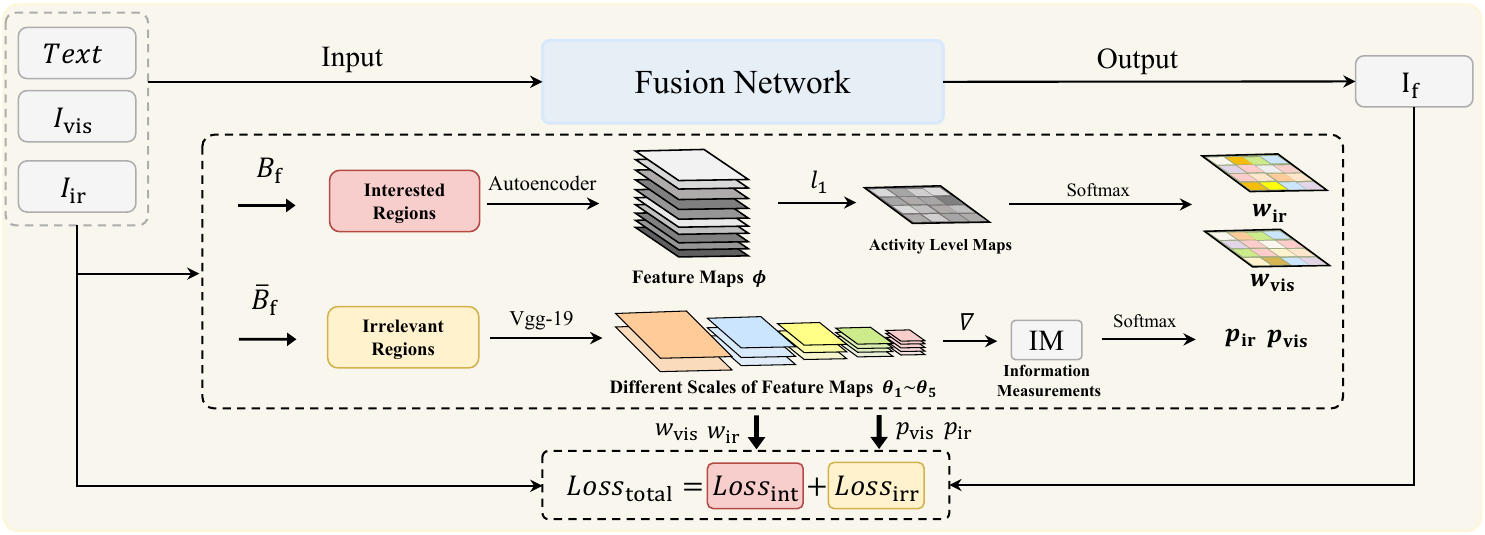}
\caption{An illustration of the training process of the proposed model.
In our implementation, we assign different levels of salience for the interested and irrelevant regions, based on activity level maps and information measurement, to obtain the weights for the input images.
Accordingly, a fusion network is optimized by using a loss function reflecting the salience differences, with the input text information used to guide the fusion process.
}
\label{figure_iilustration_training}
\end{figure*}

Finally, we add the relevant instance segmentation results together to generate the final interest map $B_{\mathrm{f}}$.
Here, an instance $O_i$ is considered relevant to the interest map if the corresponding overlapping ratio is greater than $\alpha$.
Hence, the final interest map is defined as:
\begin{equation}
    B_{\mathrm{f}} = \sum_{1\leq j\leq Q, r_j>\alpha}O_j.
\label{eq_B_final}
\end{equation}
A discussion about the threshold $\alpha$ will be presented in Section.~\ref{sectionImpactOfThreshold}.

In this way, we establish a fine-grained connection between the input image pairs and the text modality.

\subsubsection{Training}
As shown in Fig.~\ref{figure_iilustration_training}, during the training process, our loss function is designed to focus on the final interest map $B_{\mathrm{f}}$.
The total loss is formulated as
\begin{equation}
    Loss_{\mathrm{total}} = Loss_{\mathrm{int}} + Loss_{\mathrm{irr}},
\end{equation}
where $Loss_\mathrm{int}$ and $Loss_\mathrm{irr}$ stand for the region of interest and the irrelevant region loss components.
As the thermal radiation information can be captured by the pixel intensity distribution, while the texture details in the RGB images are reflected in the gradient information~\cite{ma2019fusiongan,zhang2020rethinking,zhang2021sdnet}, we use such criterion together with the principles of the dominant modality theory in the affine fusion unit to define these two loss functions.
Specifically, the weights $w(B_f, \phi(I))$ and $p(\overline{B_f},\nabla\theta(I))$ are adopted in the region of interest and irrelevant region, respectively, to measure the salience degree of source images.
$\phi(I)$ and $\theta(I)$ denote the image features obtained from pre-trained models.
$\nabla$ is the gradient operator and $\overline{B_f}(x,y) = 1-B_f(x,y)$.

Firstly, in the region of interest, we use the spatial attention strategy adopted in~\cite{yang2009multifocus,li2020nestfuse} to preserve the salient information from the infrared and visible images. 
However, different from these methods, we expect to realize the fusion process in an end-to-end manner.
To achieve this objective, we adopt the pre-trained autoencoder network~\cite{li2018densefuse} to extract the deep image features and use $\phi_i(I)$ to denote the $i$-th channel of the feature maps.
Subsequently, we conduct the $l_1$ normalisation on these features to calculate the activity level map
\begin{equation}
    A(I)(x,y) = \|\phi_{1:C}(I)(x,y)\|_1,
\end{equation}
where A denotes the activity level map and $C$ indicates the number of channels.
The activity level map can quantify the salient content of input images~\cite{yang2012pixel}.
{Since information with a higher response value is favoured by this strategy, the region of interest can be well highlighted accordingly.}
Following this, we conduct a softmax operation on $A(I_\mathrm{ir})$ and $A(I_\mathrm{vis})$ to obtain the  pixel-level weight maps $w_\mathrm{ir}$ and $w_\mathrm{vis}$.

Next, we do not follow the two-stage autoencoder-based methods.
Instead, in the region of interest, we apply these two weights in the loss function to dynamically control the proportion of pixel information transferred from the input images into the fusion result.
Thus, we obtain the formulation of this loss function
\begin{equation}
    L_\mathrm{int} = \frac{1}{H W}(L_\mathrm{int}^\mathrm{vis}+L_\mathrm{int}^\mathrm{ir}),
\end{equation}
\begin{equation}
    L_{\mathrm{int}}^\mathrm{vis} = \|B_{\mathrm{f}}\cdot w_{\mathrm{vis}}\cdot (I_{\mathrm{f}}-I_{\mathrm{vis}})\|_F^2,
\end{equation}
\begin{equation}
    L_{\mathrm{int}}^\mathrm{ir} = \|B_{\mathrm{f}}\cdot w_{\mathrm{ir}}\cdot(I_{\mathrm{f}}-I_{\mathrm{ir}})\|_F^2,
\end{equation}
where $\|\cdot\|_F$ is the Frobenius norm.

On the other hand, for the irrelevant regions, the information measurement approach~\cite{xu2020u2fusion} has been shown to handle the visual texture details well.
We tend to retain this information in the irrelevant regions as supplementary for the salient interest part.
Since we maintain significant contrast for the fused images, such difference in design can also benefit the downstream vision task to better index the position of the foreground objects. 

Specifically, we use a similar strategy to control the mixing of the source images in the non-salient areas.
Firstly, we estimate the information content of input images using a pre-trained VGG-19 network~\cite{vgg}.
This measurement is defined as
\begin{equation}
IM(I) = \frac{\sum_{i=1}^5\sum_{j=1}^{n_i}\| DS(\overline{B}_{\mathrm{f}},2^{i-1})\cdot\nabla \theta_i^j(I)\|_F^2}{n_i\sum_{i=1}^5\|DS(\overline{B}_{\mathrm{f}},2^{i-1})\|_0},
\end{equation}
where $\theta_i^j(I)$ denotes the $j$-th channel of the feature maps obtained before the $i$-th max-pooling layer of the VGG model, $n_i$ is the number of the feature maps in the corresponding convolutional layer.
$DS(X,s)$ denotes the down-sampling operation with the scale factor of $s$ on matrix $X$.

We apply a softmax operation to this information measurement to obtain the mixing weights $p_{{ir}}$  and $p_{{vis}}$.
Note that, the information measurement is calculated based on the whole input image.
Thus, different from the pixel-level weights in the area of interest, the irrelevant region loss is formulated as follows:

\begin{equation}
    L_{\mathrm{irr}} = \frac{1}{H W}( L_\mathrm{irr}^\mathrm{vis}+L_\mathrm{irr}^\mathrm{ir}),
\end{equation}
\begin{equation}
    L_\mathrm{irr}^\mathrm{vis} = p_{\mathrm{vis}}\cdot\|\overline{B}_{\mathrm{f}}\cdot(I_{\mathrm{f}}-I_{\mathrm{vis}})\|_F^2,
\end{equation}
\begin{equation}
    L_\mathrm{irr}^\mathrm{ir} = p_{\mathrm{ir}}\cdot\|\overline{B}_{\mathrm{f}}\cdot(I_{\mathrm{f}}-I_{\mathrm{ir}})\|_F^2.
\end{equation}

\begin{figure}[t]
\centering
\includegraphics[width=0.8\linewidth]{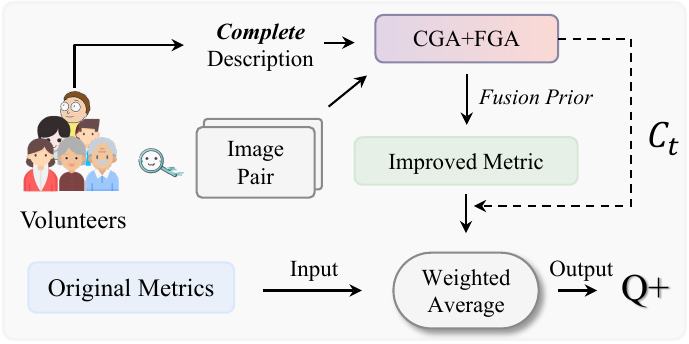}
\caption{An illustration of the textual attention assessment $Q+$. In this metric, we aim to use the fusion prior and complete description from the volunteers to improve the existing average-based fusion metrics. ($C_t$ denotes the confidence value obtained based on the coarse-to-fine association mechanism)
}
\label{figure_textual_attention}
\end{figure}

\subsection{Textual attention assessments}

Many image fusion metrics are based on common image quality assessment techniques. 
Typically, the { fusion quality measure} $Q_o(I_{\mathrm{f}},I_{\mathrm{ir}},I_{\mathrm{vis}})$ is formulated as
\begin{equation}
\label{eq_regular_metric}
   Q_o = [IQA(I_{\mathrm{f}},I_{\mathrm{ir}})+IQA(I_{\mathrm{f}},I_{\mathrm{vis}})]/2,
\end{equation}
where the image quality $IQA(X,Y)$ is defined using image $Y$ as a reference.
This approach assumes that the quality 
 of the fused image is directly related to the input images, and they are simply considered to make an equal contribution.

As shown in Fig.~\ref{figure_textual_attention}, by utilizing the textual modality to describe the region of interest within the input images, we can embed a prior fusion knowledge into the IQAs to evaluate the fusion results.
That is, depending on the purpose of the fusion task, we select salient information from the infrared modality and texture details from the visible images in the regions of interest and other areas, respectively, to construct a text-guided fused image $I_{\mathrm{tf}}$.
We generate this image in a straightforward manner:
\begin{equation}
    I_{\mathrm{tf}}=I_{\mathrm{ir}}\cdot B_{\mathrm{f}}+I_{\mathrm{vis}}\cdot\overline{B}_{\mathrm{f}}.
\end{equation}

Regarding this coarse-grained GT image as a reference, we propose a means of textual attention image fusion assessment $Q(I_{\mathrm{f}},I_{\mathrm{ir}},I_{\mathrm{vis}})^+$, which is defined as
\begin{equation}
\label{eqImprovedMetric}
    Q^+= W_\mathrm{o}\cdot Q_\mathrm{o} + C_t(I_{\mathrm{ir}},I_{\mathrm{vis}},text)\cdot IQA(I_{\mathrm{f}},I_{\mathrm{tf}}),
\end{equation}
where $W_\mathrm{o}$ and $C_t$ are the weight items and the confidence degree for the traditional and improved assessments, respectively.
$C_t$ is calculated based on the heat maps obtained by the association mechanism. It is defined as:
\begin{equation}
C_{\mathrm{t}}=\frac{\sum B_{\mathrm{f}}\cdot Max(\hat{M}(I_{\mathrm{ir}}),\hat{M}(I_{\mathrm{vis}}))}{\sum_{x=1}^{H} \sum_{y=1,\hat{B}(x,y)\neq 0}^{W} B_{\mathrm{f}}(x,y)}.
\end{equation}

The main idea is to use the average response value between the text and vision signals to indicate the extent to which the text information matches the semantics content in the source images.
During the evaluation, we impose the constraint that $W_\mathrm{O}+C_{\mathrm{t}}=1$.
In this way, all the factors, \textit{e.g.}, the purpose of the fusion task and the original image quality assessment are captured by the metrics.
The traditional evaluation indices are better adapted to the infrared and visible image fusion task. 

\section{Experimental results and analysis}\label{experiment}

\subsection{Experimental setting}
In this paper, our TextFusion is used to implement the infrared and visible image fusion task.
As there is no available infrared and visible image dataset with the corresponding text modality, we carefully annotated data from several RGBT datasets, $i.e.$, TNO, RoadScene and LLVIP~\cite{2014TNO,xu2020u2fusion,jia2021llvip}, to build a benchmark text-guided image fusion dataset (IVT).
The final database which contains $11,405$ textual annotations will be released to the research community.

During the training process, 10,000 RGBT and text triples from the IVT-LLVIP dataset were used to optimize the fusion model. 
The segmentation and the vision-and-language models are only pre-trained on the ImageNet classification tasks and the masked language modelling tasks~\cite{wu2019detectron2,kim2021vilt}.
In the test phase, the entire IVT-TNO and IVT-RoadScene datasets, along with the remaining images from the IVT-LLVIP dataset, are used to evaluate the performance.
For the quantitative comparison, some widely used metrics, \textit{i.e.}, an objective image fusion performance measure $Q_{abf}$, structural similarity (SSIM), and visual information fidelity (VIF)~\cite{xydeas2000qabf,ma2020ddcgan} are chosen to measure the similarity of the fused image and source input from different perspectives.
Besides, we also include two non-reference metrics, \textit{i.e.}, spatial frequency (SF) and standard deviation (SD) to evaluate the image quality of the fusion results.
Larger values of these metrics indicate better fusion performance.

The comparison algorithms include the mainstream paradigms that emerged in the last few years, namely a GAN-based method, TarDAL~\cite{liu2022target}, an autoencoder-based method, RFN-Nest~\cite{li2021rfn}, a self-evolution paradigm MUFusion~\cite{tang2022SeAFusion}, two methods combined with other vision tasks ReCoNet~\cite{huang2022reconet} and MetaFusion~\cite{liu2022target}, a diffusion-based model DDFM~\cite{Zhao_2023_ICCV_DDFM} and a novel representation guided network, LRRNet~\cite{li2023lrrnet}.
The experiments are conducted on an NVIDIA GeForce RTX 3090 GPU.

\begin{figure*}[t]
\centering
\includegraphics[width=1\linewidth]{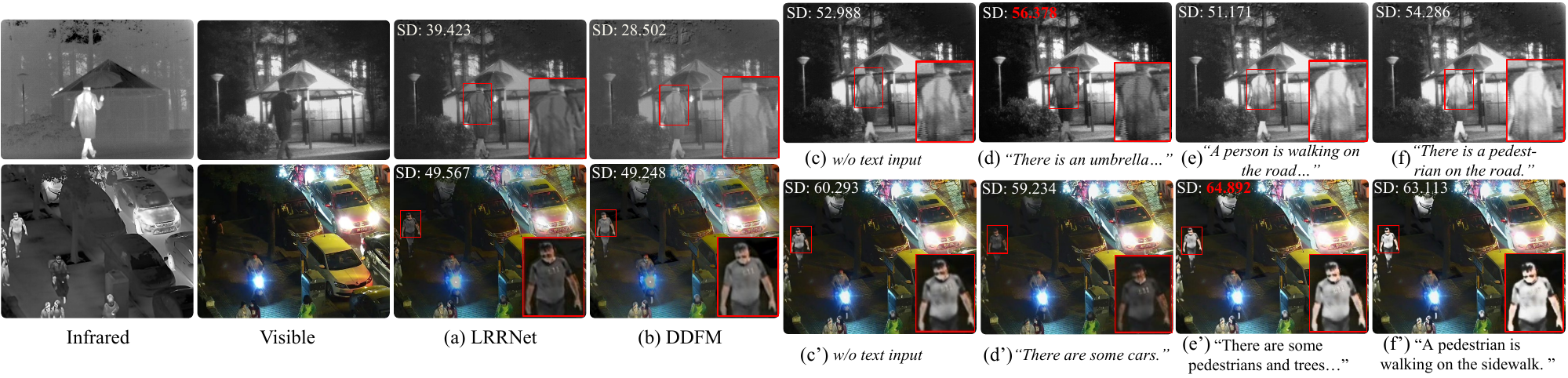}
\vspace{-2mm}
\caption{The controllability characteristics of the proposed TextFusion on two pairs of images from the IVT dataset.
While producing high-quality fusion results (reflected by SD), our method can respond to different textual descriptions.}
\vspace{-2mm}
\label{figure_customised_qualitativeTNO}
\end{figure*}

\begin{figure*}[t]
\centering
\includegraphics[width=1\linewidth]{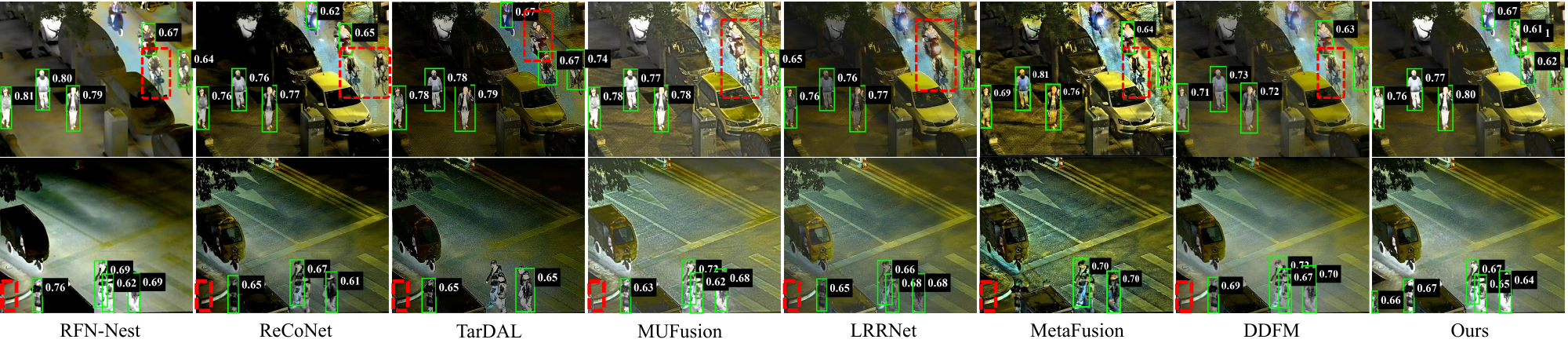}
\caption{
Pedestrian detection results obtained on images from the IVT-LLVIP dataset.
The detector can precisely locate the salient pedestrians in our fused images.
While the detection results for other modalities have false negatives  (\textcolor[rgb]{ 1,  0,  0}{red boxes}) or have lower confidence.
} 
\vspace{-3mm}
\label{figure_qualitative_detection}
\end{figure*}

\subsection{Controllability characteristics} 
In this paper, we propose a new image fusion paradigm
based on the input textual description.
The text modality enables us to obtain fusion results guided by different textual specifications of the information of interest in a given RGBT image pair.
In Fig.~\ref{figure_customised_qualitativeTNO}, we present some fused images from the advanced fusion methods as well as our TextFusion with different text inputs.

Firstly, we note that, regardless of the text input, our method consistently produces output of higher quality (SD) than that of the other two approaches.
Besides, as highlighted by the red boxes, the differences between the various regions' interests specified by the textual semantics are well reflected in the fused images.
Specifically, compared with the result without a text input, in output (e), when the complete description for the source images is added, the thermal radiation from the infrared modality is clearly preserved in the output image.
Interestingly, if we remove the word ``person" from the description, as seen from the result (d), the thermal radiation is less notable.
Furthermore, in case (f), the aim is to show that even with a simple textual input, our method has the capacity to generate fused images that are equally promising as those produced with a complete description.
Finally, as shown in output (f'), if we focus on describing the particular object, \textit{i.e.}, pedestrians, the corresponding targets can be better highlighted by the thermal information, compared to result (e').
Such differences also reveal that our TextFusion is capable of generating the desired output according to the intended semantics of the text.

\begin{table}[tbp]
  \centering
      \caption{The average detection precision of different fusion methods obtained on images from the IVT-LLVIP dataset. (\textbf{\underline{Bold}}: best, \textbf{Bold}: second best)}
  \label{QuantitativeDetection} 
  \resizebox{1\linewidth}{!}{    
    \begin{tabular}{cccccccccc}
    \hline
    Modality & RFN   & ReC   & Tar   & MUF   & LRR   & Meta  & DDF   & \cellcolor[rgb]{ .851,  .851,  .851}w/o Text & \cellcolor[rgb]{ .851,  .851,  .851}w Text \\
    \hline
    mAP@.5 & 93.6\% & 94.0\%  & 92.3\% & 94.2\% & 93.1\% & 91.4\% & 94.8\% & \cellcolor[rgb]{ .851,  .851,  .851}\textbf{95.2\%} & \cellcolor[rgb]{ .851,  .851,  .851}\underline{\textbf{95.9\%}} \\
    \hline
    \end{tabular}%
}
\vspace{-4mm}
\end{table}%

\begin{figure*}[t]
\centering
\includegraphics[width=1\linewidth]{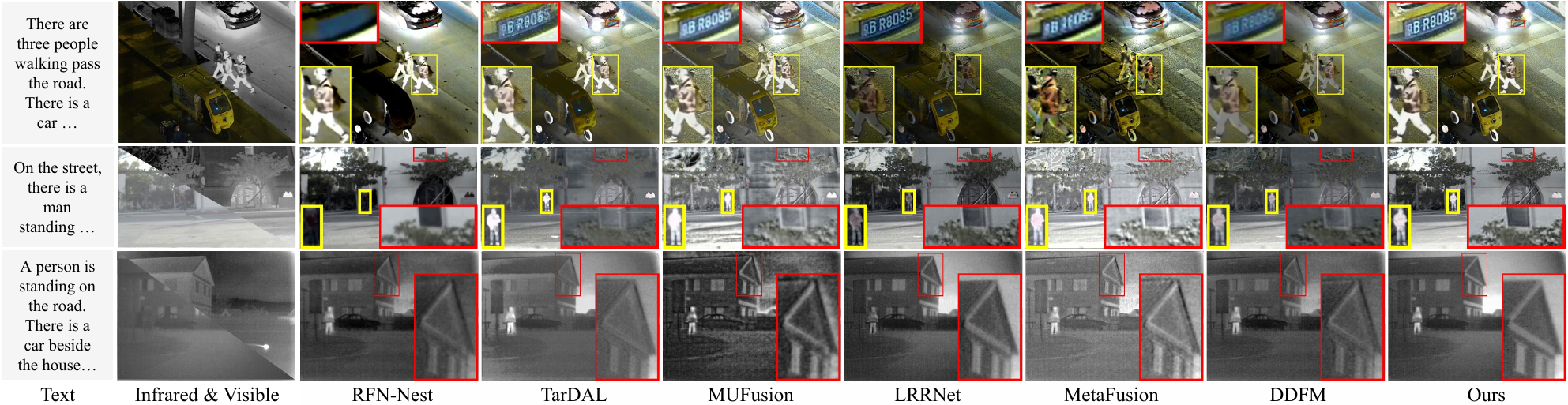}
\vspace{-2mm}
\caption{Sample qualitative results of our TextFusion compared with other state-of-the-art methods on the IVT dataset.
As shown in the highlighted regions, our method can present significant thermal information while retain clearer textual details form the RGB images, compared with other advanced approaches.
}
\label{figure_qualitativeLLVIP}
\end{figure*}

\begin{table*}[tbp]
  \centering
  \caption{Quantitative results of different image fusion methods obtained on the IVT dataset. ($mRank+$ and $mRank$ denote the mean rankings of the three textual attention and traditional metrics).}
    \label{Table_LLVIP_quantitative}  
  \resizebox{\linewidth}{!}{ 
    \begin{tabular}{cc|cccc|cccc|cc||ccc||ccc}
    \hline
    \multicolumn{12}{c||}{Dataset: LLVIP}                                                         & \multicolumn{3}{c||}{Dataset: TNO} & \multicolumn{3}{c}{Dataset: RoadScene} \\
    \hline
    Method & Venue & mRank+$\downarrow$ & Qabf+ & SSIM+ & VIF+  & mRank$\downarrow$ & Qabf  & SSIM  & VIF   & SF    & SD    & mRank+$\downarrow$ & SF    & SD    & mRank+$\downarrow$ & SF    & SD \\
    \hline
    RFN-Nest & 21' Inf. Fus. & 6.67  & 0.387  & 0.645  & 0.429  & 6.00  & 0.384  & \underline{\textbf{0.661 }} & 0.466  & 10.682  & 39.719  & 6.00  & 6.185  & 36.583  & 5.00  & 11.363  & \underline{\textbf{71.128 }} \\
    ReCoNet & 22' ECCV & 6.00  & 0.468  & 0.469  & 0.661  & 6.00  & 0.462  & 0.440  & 0.727  & 17.082  & 48.761  & 4.67  & 7.252  & 39.061  & 5.67  & 9.806  & 41.791  \\
    TarDAL & 22' CVPR & 5.67  & 0.408  & 0.552  & 0.722  & 4.67  & 0.444  & 0.598  & \textbf{0.809 } & 16.659  & \underline{\textbf{52.106 }} & 4.33  & 11.969  & 44.424  & 3.67  & 13.071  & 45.242  \\
    MUFusion & 23' Inf. Fus. & 4.33  & \underline{\textbf{0.533 }} & 0.530  & 0.716  & \textbf{4.00 } & \underline{\textbf{0.547 }} & 0.553  & 0.755  & 17.272  & 40.104  & 5.33  & 10.332  & 45.807  & 6.33  & 16.389  & 60.028  \\
    LRRNet & 23' TPAMI & 7.00  & 0.407  & 0.622  & 0.312  & 7.33  & 0.406  & 0.620  & 0.342  & 13.346  & 29.826  & 4.67  & 9.808  & 39.429  & 6.33  & 15.568  & 51.002  \\
    MetaFusion & 23' CVPR & 4.33  & 0.425  & 0.574  & \underline{\textbf{1.467 }} & 5.33  & 0.436  & 0.546  & \underline{\textbf{1.539 }} & \underline{\textbf{23.749 }} & \textbf{49.935 } & 6.33  & \underline{\textbf{15.076 }} & 47.364  & 6.00  & \underline{\textbf{22.756 }} & 48.691  \\
    DDFM  & 23' ICCV & 5.67  & 0.433  & 0.616  & 0.470  & 5.00  & 0.447  & 0.633  & 0.514  & 14.171  & 40.405  & 6.00  & 7.395  & 34.900  & 5.00  & 11.853  & 54.020  \\
    
    \rowcolor[rgb]{ .851,  .851,  .851} TextFusion w/o Text & Ours  & \textbf{3.33 } & 0.502  & \underline{\textbf{0.670 }} & 0.581  & 4.33  & 0.482  & 0.626  & 0.623  & 17.831 & 42.340  & \textbf{4.00 } & 12.324 & \textbf{48.891 } & \textbf{3.00 } & \textbf{19.212 } & \textbf{62.804 } \\
    \rowcolor[rgb]{ .851,  .851,  .851} TextFusion & Ours  & \underline{\textbf{2.00 }} & \textbf{0.527 } & \textbf{0.656 } & \textbf{0.744 } & \underline{\textbf{2.33 }} & \textbf{0.522 } & \textbf{0.642 } & 0.805  &\textbf{18.872 }   & 49.578  & \underline{\textbf{3.67 }} & \textbf{12.471 }  & \underline{\textbf{50.074 }} & \underline{\textbf{3.00 }} & 18.697  & 61.485  \\
    \hline
    \end{tabular}%
}
\vspace{-3mm}
\end{table*}%

\subsection{Pedestrian detection}
In the low-light environment, pedestrians in the infrared images contain significant thermal radiation information, which can support pedestrian-related vision tasks~\cite{jia2021llvip,zhang2021sdnet}.
The aim of our experiment is to test, how the fusion results obtained by various fusion methods help to fine-tune the YOLOv5 model on the LLVIP dataset~\cite{redmon2016yolo,jia2021llvip}.
This allows us to evaluate the performance of different fusion methods in the context of pedestrian detection tasks.

As shown in the first row of Fig.~\ref{figure_qualitative_detection}, based on our results, the detector accurately locates the salient pedestrians in the crowd.
Relying on other modalities, the detector can also identify the positions of certain pedestrians, albeit with lower confidence in their bounding box predictions compared to our method (first example).
Besides, it should be noted that in these two examples, the detector fails to recognise some of the pedestrians based on the fused images produced by other methods.
In contrast, our method exhibits the capability to accurately detect even challenging examples located in the corners (second row).
In addition, the fusion results of RFN-Nest effectively capture the salient information from the infrared modality, but they fail to preserve the texture details in the background.
This deficiency hinders their detection precision.

Apart from the results visualisation, we also conducted quantitative experiments on this task.
As presented in Table~\ref{QuantitativeDetection}, with or without text input, our TextFusion always ranks first among these advanced image fusion methods.

The above subjective and objective experiments indicate that, the proposed approach excels in effectively combining the strengths of the source images and greatly benefits downstream vision tasks.

\subsection{Image fusion results on the IVT dataset}

\textbf{Visualisation}: In Fig.~\ref{figure_qualitativeLLVIP}, we select three pairs of images from the IVT dataset to visualise the results of different image fusion methods.
As shown in the first row of this figure, the approaches to fusion adopted by these methods can be divided into two categories.
The first category, \textit{i.e.}, which includes LRRNet, MetaFusion and DDFM (last 3 methods), tends to focus more on the visible images.
The second category, comprising \textit{i.e.}  RFN-Nest, TarDAL and MUFusion, leans towards the infrared modality in order to preserve the salient thermal radiation information in the foreground region.
In contrast,  the proposed method represents a unique amalgamation of these two categories, providing a balanced combination of the salient foreground targets and the high-quality texture details present in the background.
Although the competitors adhere to the same underlying principles, by measuring only the appearance information to determine the relative weights of the source images, their fusion performance is compromised.

Specifically, as marked by the yellow boxes, compared with LRRNet and DDFM, our TextFusion contains more thermal information conveyed by the infrared image in the object regions, \textit{e.g.}, pedestrians and cars in the first example.
Although TarDAL and MUFusion also produce fused images that are comparable in these foreground regions, as evident in the fusion results of the second row, they are less effective in preserving the texture details from the visible images (red boxes) and tend to produce many unnatural artifacts around the edges of the tree.

\textbf{Quantitative results:}
For these similarity-based traditional fusion metrics (Qabf, SSIM and VIF), which belong to the standard assessment category, Eq.~(\ref{eq_regular_metric}), we use the corresponding $Metric+$ to denote the textual attention metrics.
As these traditional metrics tend to deliver different conclusions about the fusion performance~\cite{xu2020mef,li2021rfn}, \textit{i.e.}, a fusion result can have the best performance on metric A and suboptimal performance on metric B (different perspectives).
Thus, we also collect the average ranking (mRank) of these similarity-based metrics to provide insight into the fusion performance. 

As shown in Table~\ref{Table_LLVIP_quantitative}, compared to all advanced image fusion methods, our TextFusion delivers the best performance in terms of mRank on these three datasets.
This demonstrates that, the results produced by our TextFusion maintain a high correlation with input images, and that our method is able to handle different fusion scenarios robustly.
At the same time, as our method produces fused images with higher definition, it always achieves the best or second-best performance in terms of the no-reference metrics SF and SD.
This objectively measured performance is consistent with the observations made in the qualitative evaluation.

It is worth noting that, even with an empty textual input (``w/o Text"), our approach is capable of generating satisfactory results.
While the overall performance may not match that of TextFusion with text prompts, it surpasses other state-of-the-art methods.
This phenomenon indicates that our approach does not rely on elaborate text annotations.
The inherent superiority of the model design contributes to an improvement in image fusion performance.

\textbf{An analysis of the textual attention metrics:} 
With our textual attention metrics, we aim to better measure the contributions of the source images to the fusion result quality controlled with the help of linguistic semantics.
In this section, we combine the qualitative and quantitative experiments to demonstrate the effectiveness of the improved metrics.
As shown in Fig.~\ref{figure_qualitativeLLVIP}, the image fusion results of RFN-Nest suffer from the bias issue, \textit{i.e.}, the output either excessively preserves information from the RGB image or emphasizes prominent thermal radiation content at the expense of losing texture details.
However, the traditional image fusion evaluation techniques, by evenly weighting the contributions from the source images, obscure this drawback.
As a result, RFN-Nest still manages to achieve the second-best performance in terms of SSIM (Table~\ref{Table_LLVIP_quantitative}).
By contrast, our assessment (SSIM+) is able to identify this bias, with its average fusion ranking being suppressed to a certain extent (it drops from 6.00 to 6.67).
Meanwhile, as MetaFusion and LRRNet can simultaneously preserve the important content from both modalities (though with lower quality than that produced by our method), their mRank benefits from the proposed assessments.
In general, our textual attention metric can objectively identify the methods that fall short in certain aspects and promote those aligning with the fusion task objectives.
This ensures that numerical results reflect more accurately the behaviour of image fusion approaches.

%

\begin{figure}[t]
\centering
\includegraphics[width=1\linewidth]{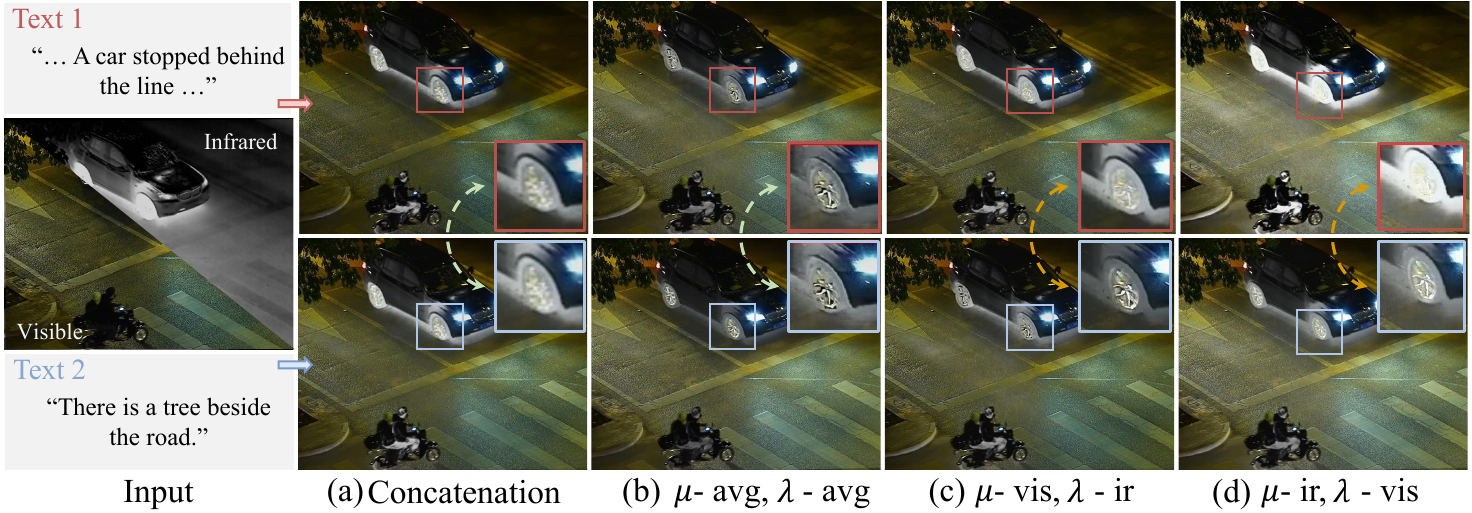}
\caption{
Qualitative results of the ablation experiments relating to the affine fusion unit with two different descriptions.}
\vspace{-3mm}
\label{figure_qualitative_affine_fusion_unit_ablation}
\end{figure}

\begin{table}[tbp]
  \centering
  \caption{Quantitative results of the ablation study relating to the affine fusion unit.}
  \vspace{-2mm}
  \label{table_ablation_affine_fusion_unit}
  \resizebox{\linewidth}{!}{ 
    \begin{tabular}{ccccccc}
    \hline
    Setting & Controllability & Qabf+ & SSIM+ & VIF+  & SF    & SD \\
    \hline
    (a) Concatenation & None  & 0.448  & 0.632  & 0.552  & 17.527  & 40.674  \\
    (b) $\mu$-avg $\lambda$-avg & None  & \textbf{0.523 } & 0.652  & \textbf{0.660 } & \underline{\textbf{19.299 }} & \textbf{47.337 } \\
    (c) $\mu$-vis $\lambda$-ir & Minor & 0.491  & \underline{\textbf{0.657 }} & 0.637  & 17.897  & 45.175  \\
    \rowcolor[rgb]{ .851,  .851,  .851} Ours ($\mu$-ir $\lambda$-vis) & \underline{\textbf{Major}} & \underline{\textbf{0.527 }} & \textbf{0.656 } & \underline{\textbf{0.744 }} & \textbf{18.872 } & \underline{\textbf{49.578 }} \\
    \hline
    \end{tabular}%
}
\vspace{-4mm}
\end{table}%

\subsection{Ablation experiments}
In this section, we present ablation experiments and visualisation results to highlight the importance of various components in our TextFusion approach.

\subsubsection{Ablation study relating to the affine fusion unit}
\label{ablationAffineFusionUnit}
To validate the effectiveness of the proposed affine fusion unit, we use 4 different feature aggregation schemes to train the TextFusion network.
In setting (a), the affine fusion unit is not used at all, \textit{i.e.}, the vision and linguistic features are simply concatenated.
The other 3 techniques are variants of the affine fusion unit, with the image modalities being used to produce the weight term $\mu$ and the bias term $\lambda$ in different ways.
Specifically, technique (b) averages the features of source images for both  $\mu$ and $\lambda$; (c) uses the RGB features for $\mu$ and the infrared features for $\lambda$; (d) uses the infrared features for $\mu$ and the RGB features for $\lambda$.

As shown in Fig.~\ref{figure_qualitative_affine_fusion_unit_ablation}, simply concatenating the linguistic and visual features along the channel dimension (setting (a)) or feeding the average vision features into the affine unit (setting (b)), our method fails to respond to different descriptions.
By incorporating fusion priors into this unit (the last two columns), specifically utilizing different modalities to determine the weight and bias terms, the text specified vision content is enabled to preserve more thermal radiation information.
The text dependent control enables the production of diverse fusion outputs.
Most importantly, our method shows enhanced salience on the targets, compared to generating the weight term using the visible light image.

As shown in Table.~\ref{table_ablation_affine_fusion_unit}, the image fusion assessments are also consistent with our observation of the qualitative results.
In particular, using the proposed setting enables our results to obtain the best performance regarding the ability of preserving the texture details (Qabf) and maintaining the information fidelity (VIF).
Besides, in terms of controllability and the imaging quality, our fusion strategy also has better performance than the tested alternatives.
It is worth noting that, when the visible modality is used to generate the weight term. 
With moderate controllability, the average performance of setting (c) on this dataset cannot catch that of setting (b).
This may due to the reason that such control manner is not consistent with our loss function design.
Consequently, it dose not well preserve salient thermal information as our TextFusion.
Furthermore, using the average operation, the original metric item (average style) in our textual assessment also contributes to the higher performance of setting (b).

\begin{figure*}[t]
\centering
\includegraphics[width=1\linewidth]{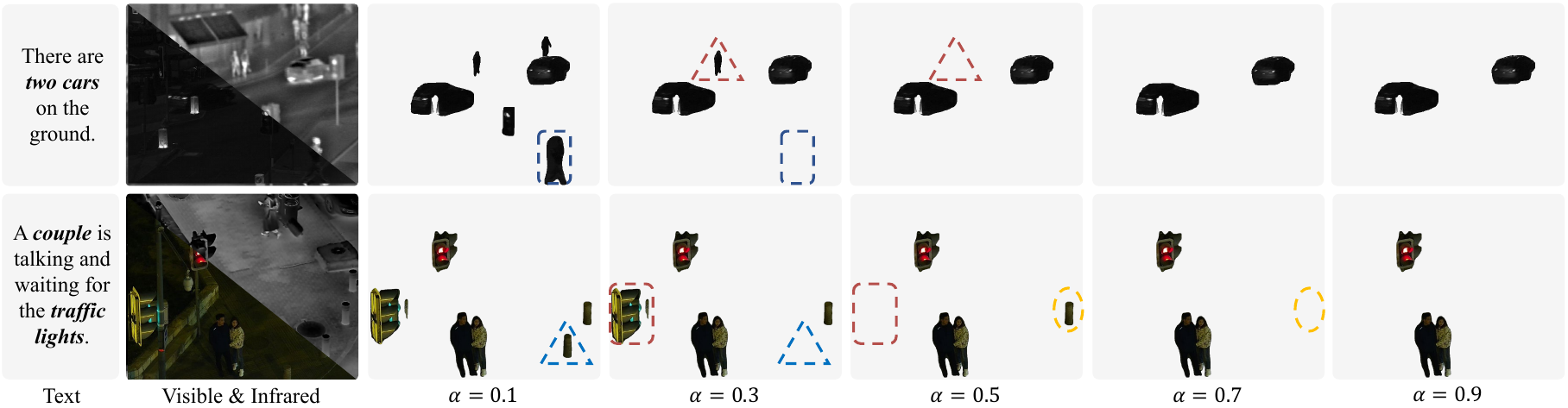}
\caption{
Visualisation results of the interested map $B_{\mathrm{f}}$ with different settings of the threshold $\alpha$.
When the hyper-parameter is set around 0.5, our pipeline can deliver more accurate association maps.
The original $B_{\mathrm{f}}$ is a binary map, we regard it as the alpha channel of the visible image to present the results.}
\label{figure_qualitative_ablationAlpha}
\end{figure*}

\subsubsection{The impact of the threshold $\alpha$ of the overlapping ratio}
\label{sectionImpactOfThreshold}
During the training process, we used the ViLT model together with the segmentation technique to obtain the fine-grained interest map.
In this section, we conduct ablation experiments to present the impact of the threshold $\alpha$ in Eq.~(3).
As shown in Fig.~\ref{figure_qualitative_ablationAlpha}, with the increase of this hyper-parameter, we can obtain a more accurate association map $B_{\mathrm{f}}$.
For example, in the first row, when the $\alpha$ is lower than 0.5, the interest maps will include some text-independent regions, \textit{e.g.}, ``pedestrians" and ``street lamp".
However, once this threshold is set higher than 0.5, the missing issues may occur in the interest maps, $i.e.$, in the second example, the green traffic light mentioned in the textual input disappears from the resulting maps.
Consequently, in our method, we set this hard threshold as 0.5, indicating that if half of the instance is associated with the linguistic information, this instance is considered part of the fine-grained interest map.

\begin{figure}[t]
\centering
\includegraphics[width=1\linewidth]{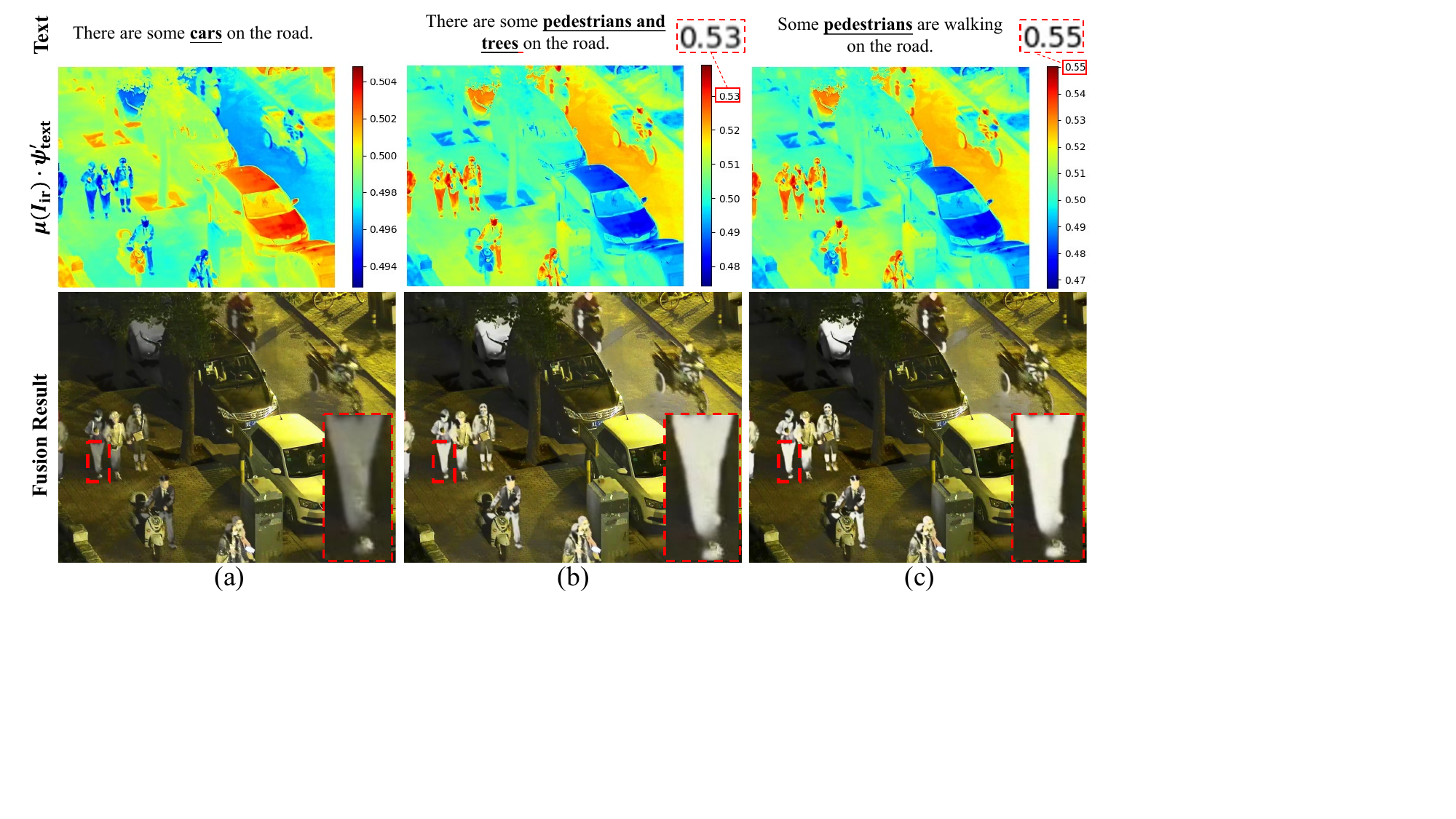}
\caption{
visualisation results of the affine fusion unit.
In the second row, we visualize the heat maps of the multiplication of the weight term and the text features (zoom in for a better view).}
\label{figure_qualitative_visualize_multiplication}
\end{figure}

\subsubsection{Visualisation of the proposed affine fusion unit}
In our TextFusion network, we design an affine fusion unit to fuse the features from different modalities.
In this section, we use three different textual inputs to demonstrate the effectiveness of the proposed unit.
As shown in the heat maps of Fig.~\ref{figure_qualitative_visualize_multiplication}, our TextFusion can highlight the foreground or background regions in the fused images according to different textual semantics (heat maps of (a) and (b)).
Additionally, as indicated by the red boxes, when we specifically focus on describing a single object, such as a pedestrian in this case, the response value in the heat map will increase.
As a result, the thermal radiation information in output (c) is more significant than that of output (a) and (b).

\begin{table}[tbp]
  \centering
      \caption{Quantitative results of the ablation study about the network architecture. (D denotes the number of the transformer blocks)}    
      \label{table_ablation_network}
  \resizebox{\linewidth}{!}{        
    \begin{tabular}{ccccccc}
    \hline
    Setting & Model Size & Qabf+ & SSIM+ & VIF+  & SF    & SD \\
    \hline
    D=2   & \textbf{0.422 } & \underline{\textbf{0.528 }} & 0.650  & \textbf{0.659 } & \textbf{18.468 } & \textbf{45.962 } \\
    D=3   & 0.514  & \textbf{0.527 } & 0.651  & 0.647  & 18.337  & 44.638  \\
    D=4   & 0.605  & 0.519  & \underline{\textbf{0.663 }} & 0.630  & 18.450  & 42.095  \\
    \rowcolor[rgb]{ .851,  .851,  .851} Ours (D=1) & \underline{\textbf{0.330 }} & 0.527  & \textbf{0.656 } & \underline{\textbf{0.744 }} & \underline{\textbf{18.872 }} & \underline{\textbf{49.578 }}\\
    \hline
    \end{tabular}%
}
\end{table}%

\subsubsection{The impact of network depth $D$ in the vision encoder}
In our TextFusion network, we use several Swin Transformer Blocks to build the vision encoders.
In this section, we conduct ablation experiments to evaluate and select the optimal network architecture.
Specifically, our TextFusion network is trained based on different settings of $D$ from the vision encoder part.
As shown in Table.~\ref{table_ablation_network}, with the increase of the network depth (volume), the fusion results tend to have better performance on the metrics of SSIM.
However, although the similarity between the input and fused images is improved, the no-reference image metrics (SF and SD) reveal that the quality and clarity of the fusion results are not promising.
In contrast, the fusion results obtained by solely utilizing one transformer block have better performance on these perspectives.
This phenomenon can be attributed to the fact that the combination of higher-level semantics conveyed through the textual input is adequate for the low-level appearance image features to produce robust fused images.

\begin{figure}[t]
\centering
\includegraphics[width=1\linewidth]{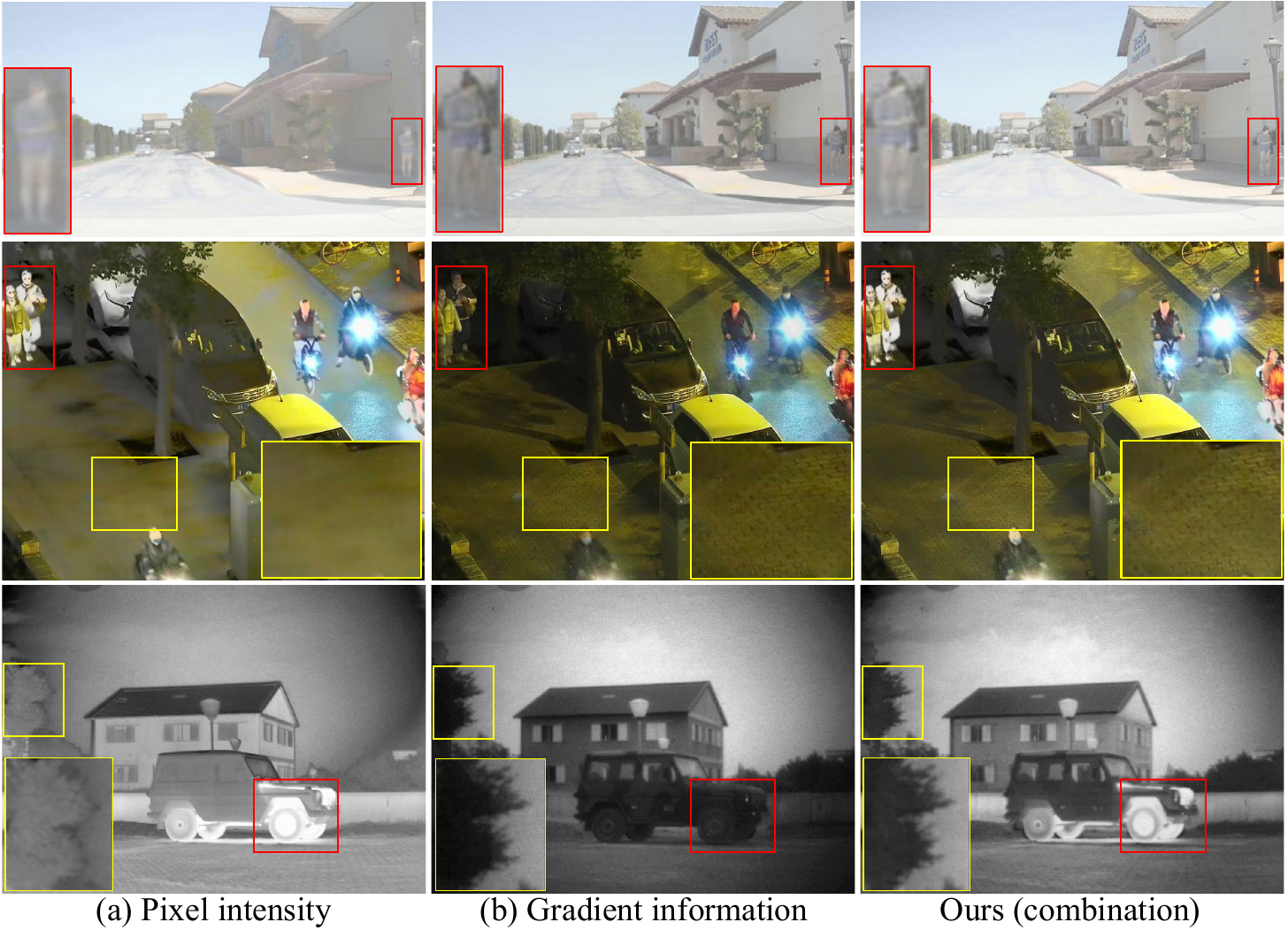}
\caption{
Qualitative results of the ablation experiments about the loss functions on 3 pairs of images from the IVT dataset.}
\label{figure_qualitative_ablation_lossfunction}
\end{figure}

\subsubsection{Effectiveness of the loss function / textual guidance}
For the network training, in order to obtain controllable fusion results guided by the textual input, different salience measures based on the activity level map (pixel intensity) and information measurement (gradient information) are used in the text-associated and irrelevant regions.
In this section, we conduct ablation experiments to demonstrate the effectiveness of the proposed loss function.
Specifically, we solely use these two measures to train the TextFusion network.
As shown in Fig.~\ref{figure_qualitative_ablation_lossfunction}, without combining these two measures, the fusion results will suffer from the issues of lacking the texture details (results (a)) or failing to preserve salient thermal radiation information (results (b)).
As our method appropriately combines these two criteria according to the high-level semantic, the significant information from both modalities, \textit{e.g.}, the salient thermal radiation information attached to the ``jeep" and the clear details of ``tree" in the third example, can be well preserved.
Such a phenomenon also demonstrates that, designing a robust fusion process relying on pure visual information is challenging.

\begin{figure}[t]
\centering
\includegraphics[width=1\linewidth]{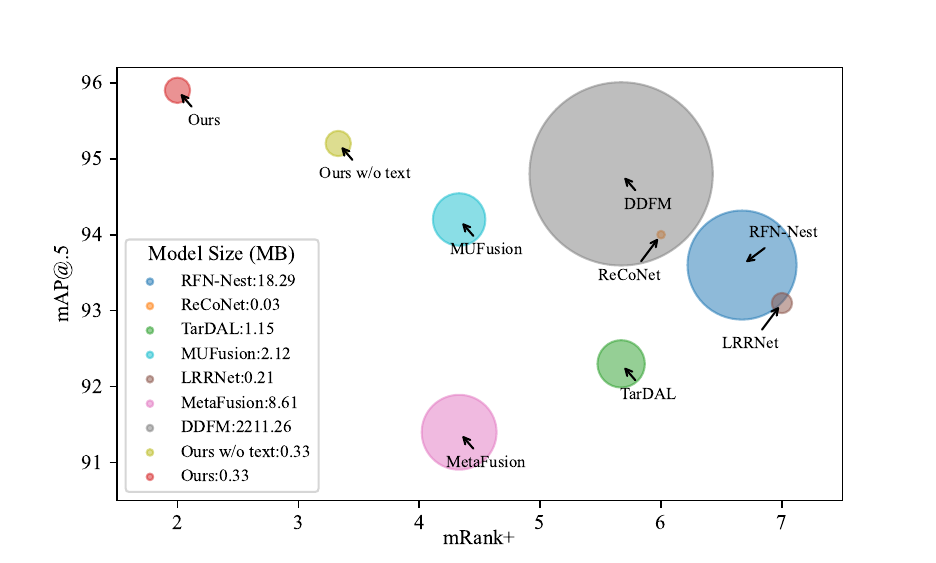}
\vspace{-4mm}
\caption{
A comparison of different methods in terms  of detection precision and average image fusion performance.
The area of the circle indicates the fusion model size of the corresponding method.
}
\vspace{-4mm}
\label{figure_model_size_map_mRank}
\end{figure}

\subsection{Model size and performance comparison}
We also conducted experiments to present the model size of different approaches and their performance on  fusion and downstream detection tasks.
As shown in Fig.~\ref{figure_model_size_map_mRank}, apart from DDFM, which is totally based on the Denoising Diffusion Probabilistic Models~\cite{ho2020DDPM}, the current image fusion models have similar model sizes when dealing with the vision modalities.
However, thanks to the introduction of textual semantics, our method gains distinct advantages and appears to be the best in all respects.

\section{Conclusion}
\label{conclusion}
In this paper, we proposed a novel paradigm for the image fusion field.
Instead of searching (by retraining) for a suitable network architecture and loss function for each image fusion use case, we introduce the textual semantics into the image fusion process to improve the mechanism of controlling and evaluating the fused images.
We demonstrated experimentally that the proposed method achieves impressive performance gains.
Apart from that, our contributions also include a text-based image quality assessment measure and the release of a benchmark dataset.

Although our TextFusion has obtained promising results, this work still has some limitations.
First of all, one of our contributions is the proposal of a new dataset, which consists of infrared and visible image pairs with textual descriptions. 
However, there is still a gap between the size of our dataset and existing benchmarks for other vision tasks.
We will continue enlarging this dataset in the future.
Secondly, in our implementation, we use the association maps to link each word and corresponding image patches.
This straightforward mechanism cannot always locate the region of interest, which may impede the fusion performance.
Lastly, our revised image fusion indexes demonstrate a greater consistency with the visualisation of different fusion results.
But the rationale behind incorporating the fusion guidance into the evaluation metrics has not been fully explored.
Research and discussion on quality metrics for image fusion is still an active and ongoing area of study.

\section{Acknowledgement}
\label{acknowledgement}
This work was supported by the National Natural Science Foundation of China (62020106012, U1836218, 62106089, 62202205), the 111 Project of Ministry of Education of China (B12018), the Engineering and Physical Sciences Research Council (EPSRC) (EP/R018456/1, EP/V002856/1), China Scholarship Council (CSC), and the Postgraduate Research \& Practice Innovation Program of Jiangsu Province (KYCX23\_2525).
 
\bibliographystyle{IEEEtran}
\bibliography{main.bib}

\begin{thebibliography}{10}
\providecommand{\url}[1]{#1}
\csname url@samestyle\endcsname
\providecommand{\newblock}{\relax}
\providecommand{\bibinfo}[2]{#2}
\providecommand{\BIBentrySTDinterwordspacing}{\spaceskip=0pt\relax}
\providecommand{\BIBentryALTinterwordstretchfactor}{4}
\providecommand{\BIBentryALTinterwordspacing}{\spaceskip=\fontdimen2\font plus
\BIBentryALTinterwordstretchfactor\fontdimen3\font minus \fontdimen4\font\relax}
\providecommand{\BIBforeignlanguage}[2]{{%
\expandafter\ifx\csname l@#1\endcsname\relax
\typeout{** WARNING: IEEEtran.bst: No hyphenation pattern has been}%
\typeout{** loaded for the language `#1'. Using the pattern for}%
\typeout{** the default language instead.}%
\else
\language=\csname l@#1\endcsname
\fi
#2}}
\providecommand{\BIBdecl}{\relax}
\BIBdecl

\bibitem{zhang2023IVIFsurveyPAMI}
X.~Zhang and Y.~Demiris, ``Visible and infrared image fusion using deep learning,'' \emph{IEEE Transactions on Pattern Analysis and Machine Intelligence}, 2023.

\bibitem{hermessi2021multimodalMedical}
H.~Hermessi, O.~Mourali, and E.~Zagrouba, ``Multimodal medical image fusion review: Theoretical background and recent advances,'' \emph{Signal Processing}, vol. 183, p. 108036, 2021.

\bibitem{zhang2021multisurveypami}
X.~Zhang, ``Deep learning-based multi-focus image fusion: A survey and a comparative study,'' \emph{IEEE Transactions on Pattern Analysis and Machine Intelligence}, 2021.

\bibitem{xu2022mefsurvey}
F.~Xu, J.~Liu, Y.~Song, H.~Sun, and X.~Wang, ``Multi-exposure image fusion techniques: A comprehensive review,'' \emph{Remote Sensing}, vol.~14, no.~3, p. 771, 2022.

\bibitem{karim2023fusionSurveyInfFus}
S.~Karim, G.~Tong, J.~Li, A.~Qadir, U.~Farooq, and Y.~Yu, ``Current advances and future perspectives of image fusion: A comprehensive review,'' \emph{Information Fusion}, vol.~90, pp. 185--217, 2023.

\bibitem{xiao2022rgbtTrackingAAAI}
Y.~Xiao, M.~Yang, C.~Li, L.~Liu, and J.~Tang, ``Attribute-based progressive fusion network for rgbt tracking,'' in \emph{Proceedings of the AAAI Conference on Artificial Intelligence}, vol.~36, no.~3, 2022, pp. 2831--2838.

\bibitem{chao2023equivalent}
J.~Chao, Z.~Zhou, H.~Gao, J.~Gong, Z.~Yang, Z.~Zeng, and L.~Dehbi, ``Equivalent transformation and dual stream network construction for mobile image super-resolution,'' in \emph{Proceedings of the IEEE/CVF Conference on Computer Vision and Pattern Recognition}, 2023, pp. 14\,102--14\,111.

\bibitem{song2023visionDehaze}
Y.~Song, Z.~He, H.~Qian, and X.~Du, ``Vision transformers for single image dehazing,'' \emph{IEEE Transactions on Image Processing}, vol.~32, pp. 1927--1941, 2023.

\bibitem{2017Multi}
Y.~Liu, X.~Chen, H.~Peng, and Z.~Wang, ``Multi-focus image fusion with a deep convolutional neural network,'' \emph{Information Fusion}, vol.~36, pp. 191--207, 2017.

\bibitem{li2018densefuse}
H.~Li and X.-J. Wu, ``Densefuse: A fusion approach to infrared and visible images,'' \emph{IEEE Transactions on Image Processing}, vol.~28, no.~5, pp. 2614--2623, 2018.

\bibitem{2021zhaoDepth_distill_mf}
F.~Zhao, W.~Zhao, H.~Lu, Y.~Liu, L.~Yao, and Y.~Liu, ``Depth-distilled multi-focus image fusion,'' \emph{IEEE Transactions on Multimedia}, pp. 1--1, 2021.

\bibitem{li2021rfn}
H.~Li, X.-J. Wu, and J.~Kittler, ``Rfn-nest: An end-to-end residual fusion network for infrared and visible images,'' \emph{Information Fusion}, vol.~73, pp. 72--86, 2021.

\bibitem{zhao2023cddfuse}
Z.~Zhao, H.~Bai, J.~Zhang, Y.~Zhang, S.~Xu, Z.~Lin, R.~Timofte, and L.~Van~Gool, ``Cddfuse: Correlation-driven dual-branch feature decomposition for multi-modality image fusion,'' in \emph{Proceedings of the IEEE/CVF Conference on Computer Vision and Pattern Recognition}, 2023, pp. 5906--5916.

\bibitem{zhang2021sdnet}
H.~Zhang and J.~Ma, ``Sdnet: A versatile squeeze-and-decomposition network for real-time image fusion,'' \emph{International Journal of Computer Vision}, pp. 1--25, 2021.

\bibitem{fu2021fuimageGAN}
Y.~Fu, X.-J. Wu, and T.~Durrani, ``Image fusion based on generative adversarial network consistent with perception,'' \emph{Information Fusion}, vol.~72, pp. 110--125, 2021.

\bibitem{zhang2020ifcnn}
Y.~Zhang, Y.~Liu, P.~Sun, H.~Yan, X.~Zhao, and L.~Zhang, ``Ifcnn: A general image fusion framework based on convolutional neural network,'' \emph{Information Fusion}, vol.~54, pp. 99--118, 2020.

\bibitem{rao2022tgfuse}
D.~Rao, X.-J. Wu, and T.~Xu, ``Tgfuse: An infrared and visible image fusion approach based on transformer and generative adversarial network,'' \emph{arXiv preprint arXiv:2201.10147}, 2022.

\bibitem{cheng2023mufusion}
C.~Cheng, T.~Xu, and X.-J. Wu, ``Mufusion: A general unsupervised image fusion network based on memory unit,'' \emph{Information Fusion}, vol.~92, pp. 80--92, 2023.

\bibitem{xu2023murf}
H.~Xu, J.~Yuan, and J.~Ma, ``Murf: Mutually reinforcing multi-modal image registration and fusion,'' \emph{IEEE Transactions on Pattern Analysis and Machine Intelligence}, 2023.

\bibitem{zhao2020realmAdaption}
F.~Zhao and W.~Zhao, ``Learning specific and general realm feature representations for image fusion,'' \emph{IEEE Transactions on Multimedia}, vol.~23, pp. 2745--2756, 2020.

\bibitem{he2023degradationICCV2023}
C.~He, K.~Li, G.~Xu, Y.~Zhang, R.~Hu, Z.~Guo, and X.~Li, ``Degradation-resistant unfolding network for heterogeneous image fusion,'' in \emph{Proceedings of the IEEE/CVF International Conference on Computer Vision}, 2023, pp. 12\,611--12\,621.

\bibitem{tang2022SeAFusion}
L.~Tang, J.~Yuan, and J.~Ma, ``Image fusion in the loop of high-level vision tasks: A semantic-aware real-time infrared and visible image fusion network,'' \emph{Information Fusion}, vol.~82, pp. 28--42, 2022.

\bibitem{liu2022target}
J.~Liu, X.~Fan, Z.~Huang, G.~Wu, R.~Liu, W.~Zhong, and Z.~Luo, ``Target-aware dual adversarial learning and a multi-scenario multi-modality benchmark to fuse infrared and visible for object detection,'' in \emph{Proceedings of the IEEE/CVF Conference on Computer Vision and Pattern Recognition}, 2022, pp. 5802--5811.

\bibitem{TANG2023divfusion}
\BIBentryALTinterwordspacing
L.~Tang, X.~Xiang, H.~Zhang, M.~Gong, and J.~Ma, ``Divfusion: Darkness-free infrared and visible image fusion,'' \emph{Information Fusion}, vol.~91, pp. 477--493, 2023. [Online]. Available: \url{https://www.sciencedirect.com/science/article/pii/S156625352200210X}
\BIBentrySTDinterwordspacing

\bibitem{Zhao2023metafusion}
W.~Zhao, S.~Xie, F.~Zhao, Y.~He, and H.~Lu, ``Metafusion: Infrared and visible image fusion via meta-feature embedding from object detection,'' in \emph{Proceedings of the IEEE/CVF Conference on Computer Vision and Pattern Recognition (CVPR)}, June 2023, pp. 13\,955--13\,965.

\bibitem{liu2023fusion_seg_ICCV2023}
J.~Liu, Z.~Liu, G.~Wu, L.~Ma, R.~Liu, W.~Zhong, Z.~Luo, and X.~Fan, ``Multi-interactive feature learning and a full-time multi-modality benchmark for image fusion and segmentation,'' in \emph{Proceedings of the IEEE/CVF International Conference on Computer Vision}, 2023, pp. 8115--8124.

\bibitem{gan2020vlp1}
Z.~Gan, Y.-C. Chen, L.~Li, C.~Zhu, Y.~Cheng, and J.~Liu, ``Large-scale adversarial training for vision-and-language representation learning,'' \emph{Advances in Neural Information Processing Systems}, vol.~33, pp. 6616--6628, 2020.

\bibitem{zhang2021vlp2}
P.~Zhang, X.~Li, X.~Hu, J.~Yang, L.~Zhang, L.~Wang, Y.~Choi, and J.~Gao, ``Vinvl: Making visual representations matter in vision-language models,'' \emph{arXiv e-prints}, pp. arXiv--2101, 2021.

\bibitem{zhou2022unsupervisedvilt}
M.~Zhou, L.~Yu, A.~Singh, M.~Wang, Z.~Yu, and N.~Zhang, ``Unsupervised vision-and-language pre-training via retrieval-based multi-granular alignment,'' in \emph{Proceedings of the IEEE/CVF Conference on Computer Vision and Pattern Recognition}, 2022, pp. 16\,485--16\,494.

\bibitem{kim2021vilt}
W.~Kim, B.~Son, and I.~Kim, ``Vilt: Vision-and-language transformer without convolution or region supervision,'' in \emph{International Conference on Machine Learning}.\hskip 1em plus 0.5em minus 0.4em\relax PMLR, 2021, pp. 5583--5594.

\bibitem{li2020manigan}
B.~Li, X.~Qi, T.~Lukasiewicz, and P.~H. Torr, ``Manigan: Text-guided image manipulation,'' in \emph{Proceedings of the IEEE/CVF Conference on Computer Vision and Pattern Recognition}, 2020, pp. 7880--7889.

\bibitem{li2022languageSegmentation}
B.~Li, K.~Q. Weinberger, S.~Belongie, V.~Koltun, and R.~Ranftl, ``Language-driven semantic segmentation,'' \emph{arXiv preprint arXiv:2201.03546}, 2022.

\bibitem{luddecke2022textImageSegmentation}
T.~L{\"u}ddecke and A.~Ecker, ``Image segmentation using text and image prompts,'' in \emph{Proceedings of the IEEE/CVF Conference on Computer Vision and Pattern Recognition}, 2022, pp. 7086--7096.

\bibitem{lin2023clipSeg}
Y.~Lin, M.~Chen, W.~Wang, B.~Wu, K.~Li, B.~Lin, H.~Liu, and X.~He, ``Clip is also an efficient segmenter: A text-driven approach for weakly supervised semantic segmentation,'' in \emph{Proceedings of the IEEE/CVF Conference on Computer Vision and Pattern Recognition}, 2023, pp. 15\,305--15\,314.

\bibitem{li2023ovtrack}
S.~Li, T.~Fischer, L.~Ke, H.~Ding, M.~Danelljan, and F.~Yu, ``Ovtrack: Open-vocabulary multiple object tracking,'' in \emph{Proceedings of the IEEE/CVF Conference on Computer Vision and Pattern Recognition}, 2023, pp. 5567--5577.

\bibitem{Liu_2021SwinTransformer}
Z.~Liu, Y.~Lin, Y.~Cao, H.~Hu, Y.~Wei, Z.~Zhang, S.~Lin, and B.~Guo, ``Swin transformer: Hierarchical vision transformer using shifted windows,'' in \emph{Proceedings of the IEEE/CVF International Conference on Computer Vision (ICCV)}, October 2021, pp. 10\,012--10\,022.

\bibitem{liu2020bilevel}
R.~Liu, J.~Liu, Z.~Jiang, X.~Fan, and Z.~Luo, ``A bilevel integrated model with data-driven layer ensemble for multi-modality image fusion,'' \emph{IEEE Transactions on Image Processing}, vol.~30, pp. 1261--1274, 2020.

\bibitem{huang2020pixelBert}
Z.~Huang, Z.~Zeng, B.~Liu, D.~Fu, and J.~Fu, ``Pixel-bert: Aligning image pixels with text by deep multi-modal transformers,'' \emph{arXiv preprint arXiv:2004.00849}, 2020.

\bibitem{ma2019fusiongan}
J.~Ma, W.~Yu, P.~Liang, C.~Li, and J.~Jiang, ``Fusiongan: A generative adversarial network for infrared and visible image fusion,'' \emph{Information Fusion}, vol.~48, pp. 11--26, 2019.

\bibitem{zhang2020rethinking}
H.~Zhang, H.~Xu, Y.~Xiao, X.~Guo, and J.~Ma, ``Rethinking the image fusion: A fast unified image fusion network based on proportional maintenance of gradient and intensity,'' in \emph{Proceedings of the AAAI Conference on Artificial Intelligence}, vol.~34, no.~07, 2020, pp. 12\,797--12\,804.

\bibitem{yang2009multifocus}
B.~Yang and S.~Li, ``Multifocus image fusion and restoration with sparse representation,'' \emph{IEEE Transactions on Instrumentation and Measurement}, vol.~59, no.~4, pp. 884--892, 2009.

\bibitem{li2020nestfuse}
H.~Li, X.-J. Wu, and T.~Durrani, ``{NestFuse: An Infrared and Visible Image Fusion Architecture based on Nest Connection and Spatial/Channel Attention Models},'' \emph{IEEE Transactions on Instrumentation and Measurement}, vol.~69, no.~12, pp. 9645--9656, 2020.

\bibitem{yang2012pixel}
B.~Yang and S.~Li, ``Pixel-level image fusion with simultaneous orthogonal matching pursuit,'' \emph{Information fusion}, vol.~13, no.~1, pp. 10--19, 2012.

\bibitem{xu2020u2fusion}
H.~Xu, J.~Ma, J.~Jiang, X.~Guo, and H.~Ling, ``U2fusion: A unified unsupervised image fusion network,'' \emph{IEEE Transactions on Pattern Analysis and Machine Intelligence}, 2020.

\bibitem{vgg}
K.~Simonyan and A.~Zisserman, ``Very deep convolutional networks for large-scale image recognition,'' in \emph{International Conference on Learning Representations}, May 2015.

\bibitem{2014TNO}
A.~Toet \emph{et~al.}, ``Tno image fusion dataset,'' \emph{Figshare. data}, 2014.

\bibitem{jia2021llvip}
X.~Jia, C.~Zhu, M.~Li, W.~Tang, and W.~Zhou, ``Llvip: A visible-infrared paired dataset for low-light vision,'' in \emph{Proceedings of the IEEE/CVF International Conference on Computer Vision}, 2021, pp. 3496--3504.

\bibitem{wu2019detectron2}
Y.~Wu, A.~Kirillov, F.~Massa, W.-Y. Lo, and R.~Girshick, ``Detectron2,'' \url{https://github.com/facebookresearch/detectron2}, 2019.

\bibitem{xydeas2000qabf}
C.~Xydeas and V.~Petrovic, ``Objective image fusion performance measure,'' \emph{Electronics letters}, vol.~36, no.~4, pp. 308--309, 2000.

\bibitem{ma2020ddcgan}
J.~Ma, H.~Xu, J.~Jiang, X.~Mei, and X.-P. Zhang, ``Ddcgan: A dual-discriminator conditional generative adversarial network for multi-resolution image fusion,'' \emph{IEEE Transactions on Image Processing}, vol.~29, pp. 4980--4995, 2020.

\bibitem{huang2022reconet}
Z.~Huang, J.~Liu, X.~Fan, R.~Liu, W.~Zhong, and Z.~Luo, ``Reconet: Recurrent correction network for fast and efficient multi-modality image fusion,'' in \emph{Computer Vision--ECCV 2022: 17th European Conference, Tel Aviv, Israel, October 23--27, 2022, Proceedings, Part XVIII}.\hskip 1em plus 0.5em minus 0.4em\relax Springer, 2022, pp. 539--555.

\bibitem{Zhao_2023_ICCV_DDFM}
Z.~Zhao, H.~Bai, Y.~Zhu, J.~Zhang, S.~Xu, Y.~Zhang, K.~Zhang, D.~Meng, R.~Timofte, and L.~Van~Gool, ``Ddfm: Denoising diffusion model for multi-modality image fusion,'' in \emph{Proceedings of the IEEE/CVF International Conference on Computer Vision (ICCV)}, October 2023, pp. 8082--8093.

\bibitem{li2023lrrnet}
H.~Li, T.~Xu, X.-J. Wu, J.~Lu, and J.~Kittler, ``Lrrnet: A novel representation learning guided fusion network for infrared and visible images,'' \emph{IEEE Transactions on Pattern Analysis and Machine Intelligence}, 2023.

\bibitem{redmon2016yolo}
J.~Redmon, S.~Divvala, R.~Girshick, and A.~Farhadi, ``You only look once: Unified, real-time object detection,'' in \emph{Proceedings of the IEEE conference on computer vision and pattern recognition}, 2016, pp. 779--788.

\bibitem{xu2020mef}
H.~Xu, J.~Ma, and X.-P. Zhang, ``Mef-gan: multi-exposure image fusion via generative adversarial networks,'' \emph{IEEE Transactions on Image Processing}, vol.~29, pp. 7203--7216, 2020.

\bibitem{ho2020DDPM}
J.~Ho, A.~Jain, and P.~Abbeel, ``Denoising diffusion probabilistic models,'' \emph{Advances in neural information processing systems}, vol.~33, pp. 6840--6851, 2020.

\end{thebibliography}
%





\begin{IEEEbiography}[{\includegraphics[width=1in,height=1.25in,clip,keepaspectratio]{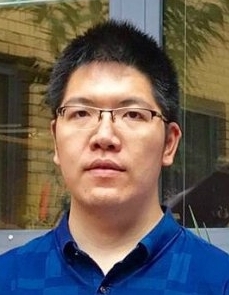}}]{Tianyang Xu} received the B.Sc. degree in electronic science and engineering from Nanjing University, Nanjing, China, in 2011. He received the PhD degree at the School of Artificial Intelligence and Computer Science, Jiangnan University, Wuxi, China, in 2019. 
He is currently an Associate Professor at the School of Artificial Intelligence and Computer Science, Jiangnan University, Wuxi, China.
His research interests include visual tracking and deep learning. 
He has published several scientific papers, including IJCV, ICCV, TIP, TIFS, TKDE, TMM, TCSVT etc. He achieved top 1 performance in academic competitions, including the VOT2018 public dataset (ECCV18), VOT2020 RGBT challenge (ECCV20), Anti-UAV challenge (CVPR20), Multi-Modal Video Reasoning and Analysing Competition (ICCV21).
\end{IEEEbiography}

\begin{IEEEbiography}[{\includegraphics[width=1in,height=1.25in,clip,keepaspectratio]{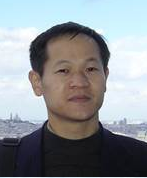}}]{Xiao-Jun Wu}
received his B.S. degree in mathematics from Nanjing
Normal University, Nanjing, PR China in 1991 and M.S. degree in
1996, and Ph.D. degree in Pattern Recognition and Intelligent System
in 2002, both from Nanjing University of Science and Technology,
Nanjing, PR China, respectively. He was a fellow of United Nations
University, International Institute for Software Technology (UNU/IIST)
from 1999 to 2000. From 1996 to 2006, he taught in the School of
Electronics and Information, Jiangsu University of Science and Technology where he
was an exceptionally promoted professor. He joined Jiangnan University in 2006 where
he is currently Dean of the Graduate School and a distinguished professor in the School of Artificial Intelligence and
Computer Science, Jiangnan University. He won the most outstanding postgraduate
award by Nanjing University of Science and Technology. He has published more than
400 papers in his fields of research. He was a visiting postdoctoral researcher in the
Centre for Vision, Speech, and Signal Processing (CVSSP), University of Surrey, UK
from 2003 to 2004, under the supervision of Professor Josef Kittler. His current research
interests are pattern recognition, computer vision, fuzzy systems, and neural networks.
He owned several domestic and international awards because of his research
achievements. Currently, he is a Fellow of the International Association for Pattern Recognition (IAPR) and the Asia-Pacific Artificial Intelligence Association (AAIA).

\end{IEEEbiography}

\begin{IEEEbiography}[{\includegraphics[width=1in,height=1.25in,clip,keepaspectratio]{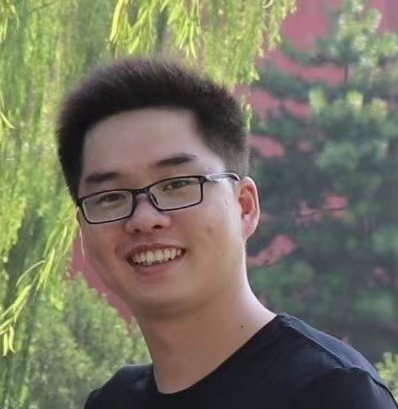}}]{Hui Li}
received the B.Sc. degree in School of Internet of Things Engineering from Jiangnan University, China, in 2015. He received the PhD degree at the School of Internet of Things Engineering, Jiangnan University, Wuxi, China, in 2022. He is currently a Lecturer at the School of Artificial Intelligence and Computer Science, Jiangnan University, Wuxi, China. His research interests include image fusion and multi-modal visual information processing.

He has published several scientific papers, including IEEE TPAMI, Information Fusion, IEEE TIP, IEEE TCYB, IEEE TIM, ICPR etc. He achieved top tracking performance in several competitions, including the VOT2020 RGBT challenge (ECCV20) and Anti-UAV challenge (ICCV21).

\end{IEEEbiography}

\begin{IEEEbiography}[{\includegraphics[width=1in,height=1.25in,clip,keepaspectratio]{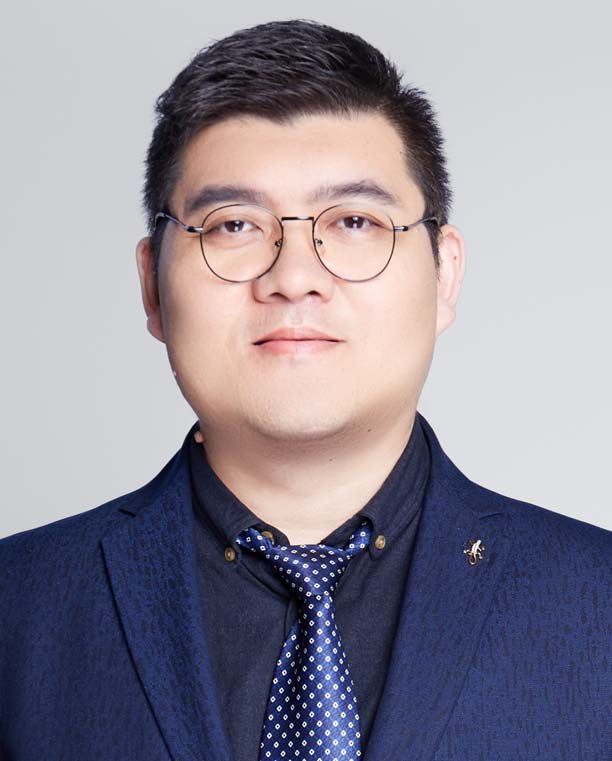}}]{Xi Li} (Senior Member, IEEE) received the Ph.D.
degree from the National Laboratory of Pattern
Recognition, Chinese Academy of Sciences, Beijing,
China, in 2009.
From 2009 to 2010, he was a Post-Doctoral
Researcher with the Centre National de la Recherche
Scientifique (CNRS) Telecom ParisTech, Paris,
France. He was a Senior Researcher with The University of Adelaide, Adelaide, SA, Australia. He is
currently a Full Professor with Zhejiang University,
Hangzhou, China. His research interests include
visual tracking, compact learning, motion analysis, face recognition, data
mining, and image retrieval.
\end{IEEEbiography}

\begin{IEEEbiography}[{\includegraphics[width=1in,height=1.25in,clip,keepaspectratio]{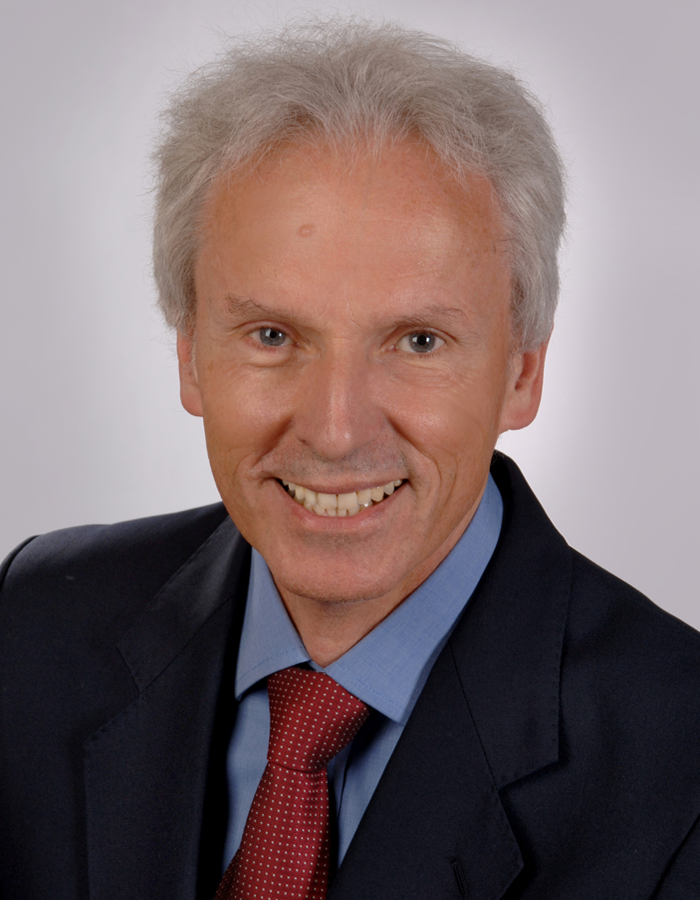}}]{Josef Kittler} received the B.A., Ph.D., and D.Sc. degrees from the University of Cambridge, in 1971, 1974, and 1991, respectively. He is a distinguished Professor of Machine Intelligence at the Centre for Vision, Speech and Signal Processing, University of Surrey, Guildford, U.K. He conducts research in biometrics, video and image database retrieval, medical image analysis, and cognitive vision. He published the textbook Pattern Recognition: A Statistical Approach and over 700 scientific papers. His publications have been cited more than 60,000 times (Google Scholar).

He is series editor of Springer Lecture Notes on Computer Science. He currently serves on the Editorial Boards of Pattern Recognition Letters, Pattern Recognition and Artificial Intelligence, Pattern Analysis and Applications. He also served as a member of the Editorial Board of IEEE Transactions on Pattern Analysis and Machine Intelligence during 1982-1985. He served on the Governing Board of the International Association for Pattern Recognition (IAPR) as one of the two British representatives during the period 1982-2005, President of the IAPR during 1994-1996.
\end{IEEEbiography}

\begin{IEEEbiography}[{\includegraphics[width=1in,height=1.25in,clip,keepaspectratio]{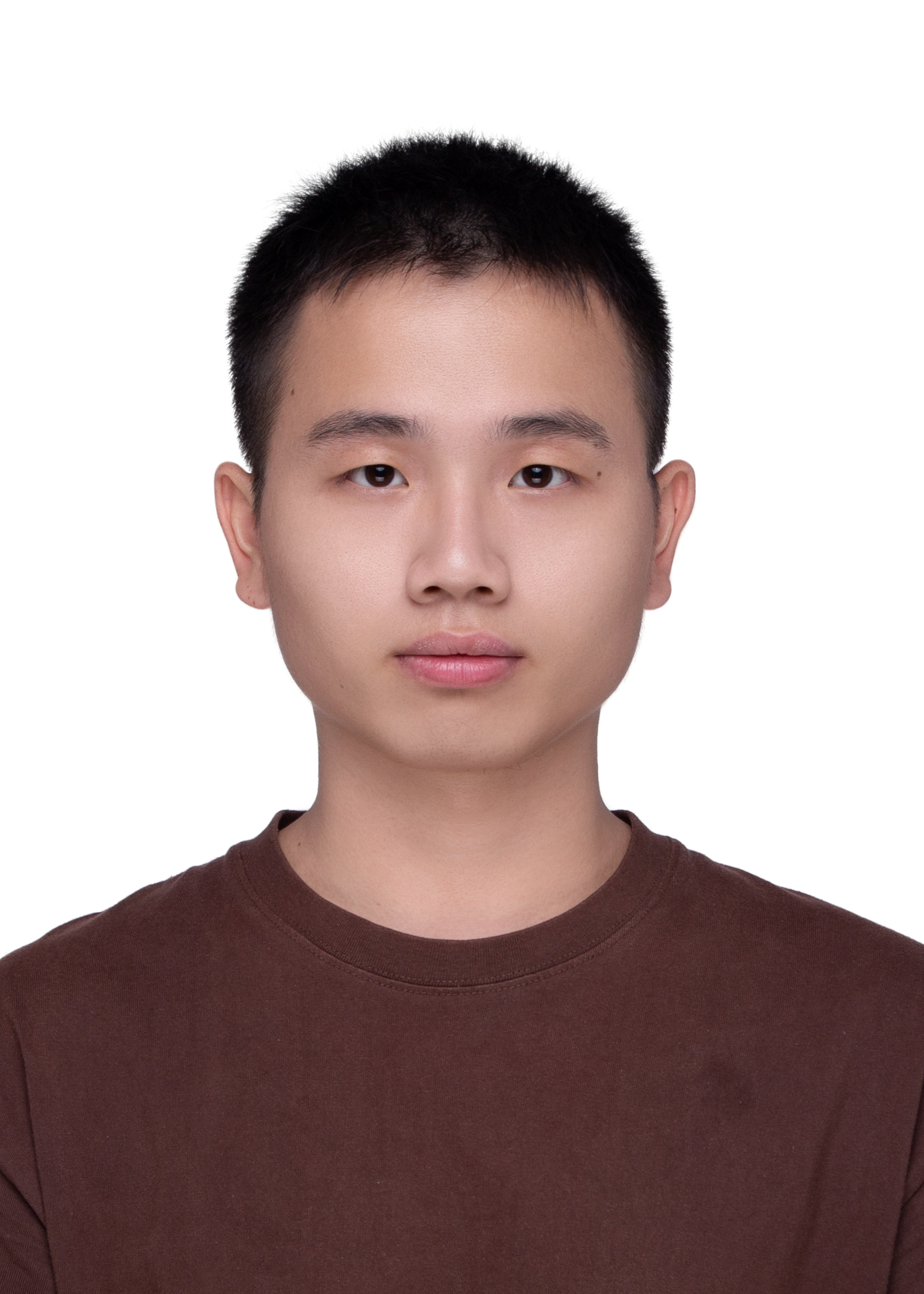}}]{Chunyang Cheng} is working toward the Ph.D degree at Jiangsu Provincial Engineering Laboratory of Pattern Recognition and Computational Intelligence, Jiangnan University, Wuxi, China. His research interests include image fusion and deep learning.
\end{IEEEbiography}

\vfill

\end{document}